%% file: main.tex
\documentclass[10pt,twocolumn,letterpaper]{article}

\usepackage{iccv}
\usepackage{times}
\usepackage{epsfig}
\usepackage{graphicx}
\usepackage{enumitem}
\usepackage{subcaption}
\usepackage{mathtools}
\usepackage{amsmath}
\usepackage{amssymb}

\newcommand{\z}{\mathbf{z}}
\usepackage[dvipsnames]{xcolor}
\usepackage{booktabs}
\usepackage{arydshln}
\usepackage{multirow}
\usepackage{lipsum}
\usepackage[pagebackref=true,breaklinks=true,letterpaper=true,colorlinks,bookmarks=false]{hyperref}

\iccvfinalcopy % *** Uncomment this line for the final submission

\def\iccvPaperID{10664} % *** Enter the ICCV Paper ID here

\ificcvfinal\fi

\begin{document}

%%%%%%%%% TITLE
\title{ContraNeRF: 3D-Aware Generative Model via Contrastive Learning \\
with Unsupervised Implicit Pose Embedding}

\author{Mijeong Kim\textsuperscript{\normalfont 1}\thanks{}\qquad \qquad Hyunjoon Lee\textsuperscript{\normalfont 3} \qquad \qquad Bohyung Han\textsuperscript{\normalfont 1,2} \\
{\hspace{-0.8cm} \textsuperscript{1}ECE \& \textsuperscript{2}IPAI, Seoul National University}~~~~~~\textsuperscript{3}Kakao Brain \\
{\tt\small \{mijeong.kim, bhhan\}@snu.ac.kr \quad malfo.lee@kakaobrain.com}
}

\ificcvfinal\thispagestyle{empty}\fi

\twocolumn[{
\maketitle
\centering
\includegraphics[width=0.98\linewidth]{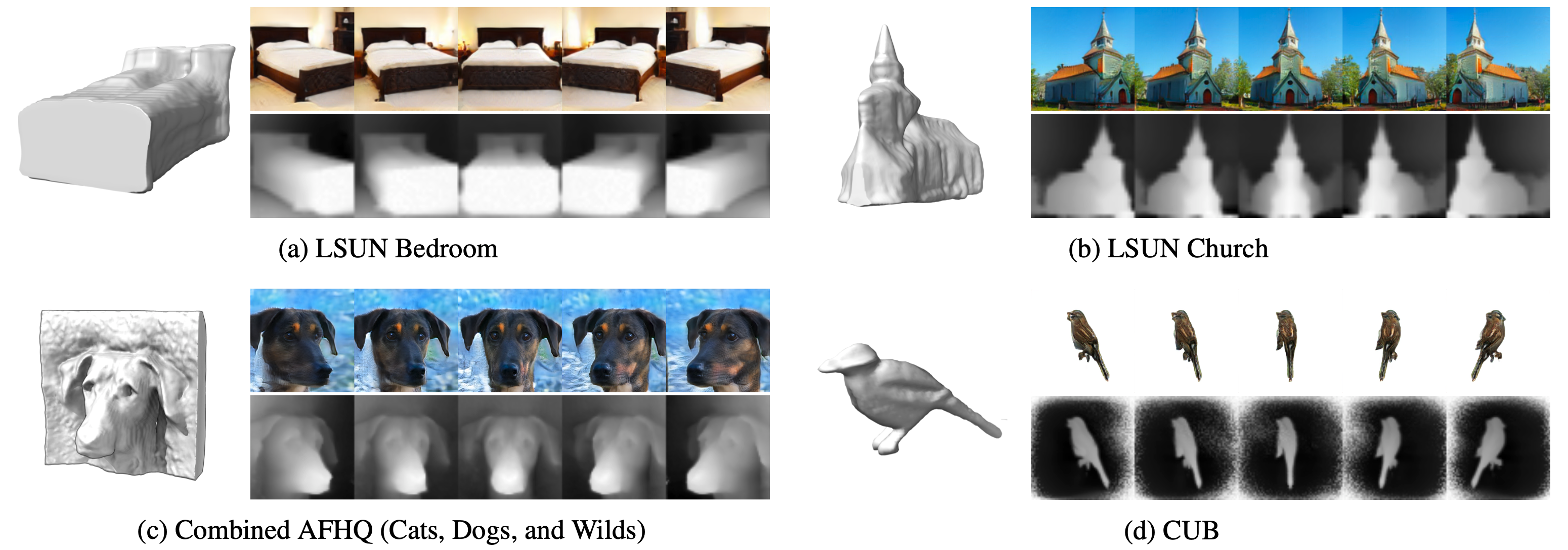}
\vspace{-2mm}
\captionof{figure}
{
Illustration of generated examples. 
    Our 3D GAN enables synthesis of complex geometric scenes such as bedroom, church, animal faces, or birds, beyond simple geometries such as human face.
Our approach trains from a collection of 2D images without ground-truth camera poses, depth information, target-specific shape priors, or multi-view supervision.
}
\label{fig:teaser_results}
\vspace{1cm}
}]
{
  \renewcommand{\thefootnote}%
    {\fnsymbol{footnote}}
  \footnotetext[1]{This work was partly done during an internship at Kakao Brain.}
}

%%%%%%%%% ABSTRACT
\input{./sections/_0_abstract}
%%%%%%%%% BODY TEXT
\input{./sections/_1_introduction}
\input{./sections/_2_related_work}
\input{./sections/_3_method}

\input{./sections/_4_experiment}
\input{./sections/_5_conclusion}
{\small
\bibliographystyle{ieee_fullname}
\bibliography{egbib}
}
\input{./sections/_supple}

\end{document}

%% file: sections/_0_abstract.tex
% !TEX root = ../main.tex
\begin{abstract}
Although 3D-aware GANs based on neural radiance fields have achieved competitive performance, their applicability is still limited to objects or scenes with the ground-truths or prediction models for clearly defined canonical camera poses.
To extend the scope of applicable datasets, we propose a novel 3D-aware GAN optimization technique through contrastive learning with implicit pose embeddings.
To this end, we first revise the discriminator design and remove dependency on ground-truth camera poses.
Then, to capture complex and challenging 3D scene structures more effectively, we make the discriminator estimate a high-dimensional implicit pose embedding from a given image and perform contrastive learning on the pose embedding.
The proposed approach can be employed for the dataset, where the canonical camera pose is ill-defined because it does not look up or estimate camera poses.
Experimental results show that our algorithm outperforms existing methods by large margins on the datasets with multiple object categories and inconsistent canonical camera poses.
\end{abstract}

%% file: sections/_1_introduction.tex
% !TEX root = ../main.tex
\section{Introduction}

\begin{figure}[t]
    \center
    \includegraphics[trim={0 0 0 0},clip,width=0.99\linewidth]{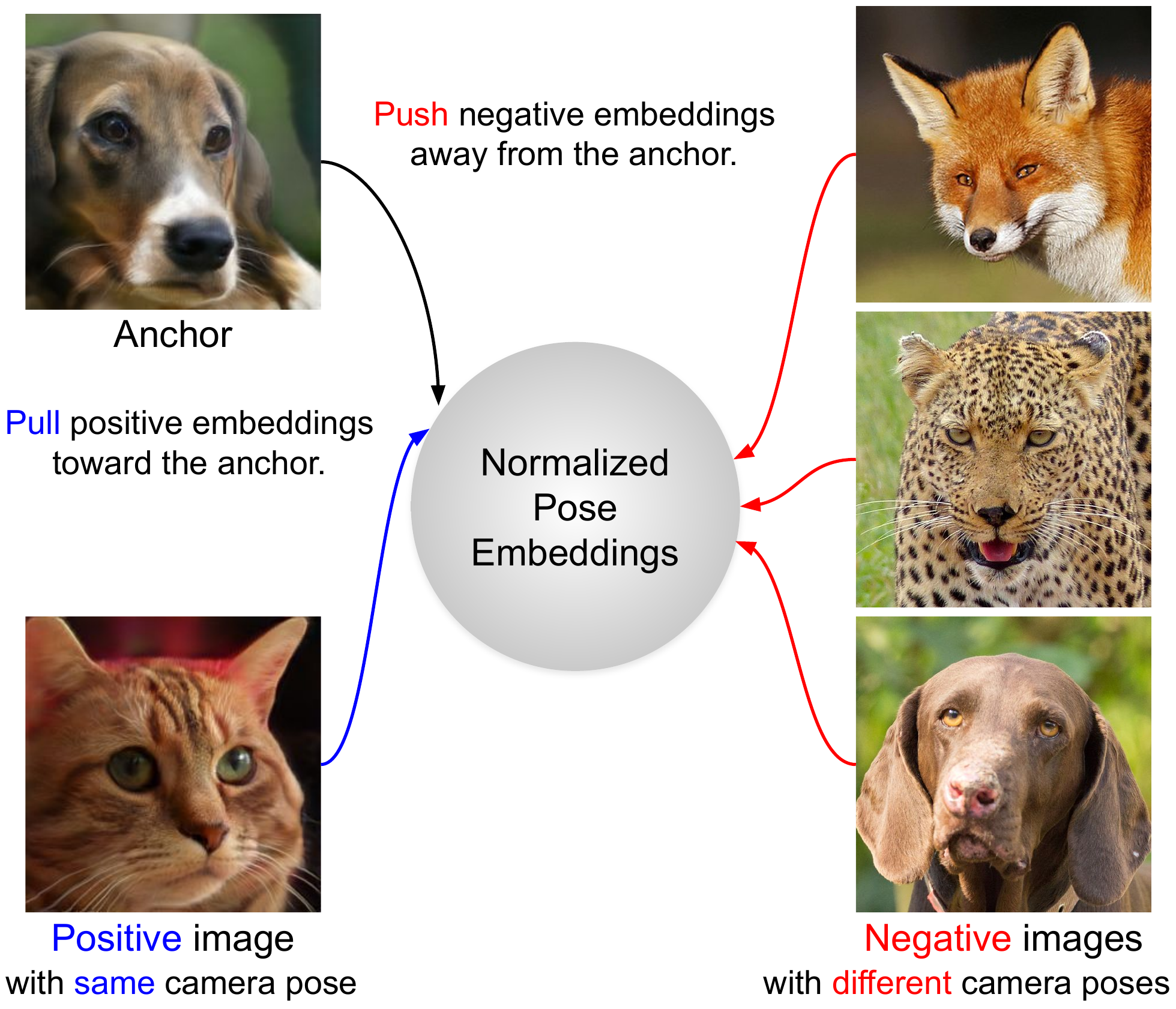}
    \caption{Illustration of the proposed contrastive learning on the pose embedding space.
	The `positive' and `negative' images denote images rendered in the same or different directions with the `anchor' image, respectively.
    The distance between pose embeddings of positive pairs are learned to be closer than those of negative pairs.
    }
    \label{fig:teaser}
    \vspace{-2mm}
\end{figure}
3D-aware Generative Adversarial Networks (GANs) aim to synthesize multiple views of a single scene with explicitly control of camera poses.
Recent methods~\cite{chan2022efficient,cai2022pix2nerf,niemeyer2021campari,chan2021pi,pan2021shading,or2022stylesdf,sun2022fenerf,deng2022gram,gu2021stylenerf,schwarz2020graf,skorokhodov2022epigraf} incorporate the advances of neural radiance fields~\cite{park2019deepsdf,niemeyer2020differentiable,mildenhall2021nerf} into generative models~\cite{goodfellow2014generative,karras2019style,karras2020training} and reconstruct 3D scenes using a collection of 2D images without 3D shape priors.
This technique allows us to predict not only 2D projected images but also their underlying 3D structures.
However, they still suffer from the limited scope of target domains; most algorithms deal with only a few object categories, \eg, human or cat faces, where ground-truth or estimated camera poses are available and the canonical camera pose is well-defined.
Although there exist some methods~\cite{li2022depthgan, devries2021gsn, bautista2022gaudi} that extend the scope of target domains to realistic ones with less geometric priors, they rely on additional geometric cues such as depth maps of training examples.

To alleviate the drawbacks of existing 3D-aware GAN approaches, we design a novel discriminator for representing the 3D structure of complex scenes in diverse domains without extra information.
As an intermediate goal of our algorithm, we start from removing the dependency on ground-truth camera poses in the discriminator employed in previous work~\cite{chan2022efficient} and make the discriminator learn a camera pose regressor in a self-supervised way using generated images and their rendering poses.
Such a simple approach turns out to be effective in synthesizing novel views without ground-truth camera poses of training images, but it sometimes fails to reconstruct 3D structures properly.
We argue that the limitation is mainly due to explicit camera pose regression, which is not desirable to handle scenes or objects with complex and heterogeneous geometries.

To further improve the quality of complex geometric and photometric structures, we propose implicit camera pose embedding in a high-dimensional space for robust and comprehensive 3D reconstruction.
For training, we employ a self-supervised contrastive learning~\cite{oord2018representation}, which captures rich geometric information of scenes from diverse pairwise relations of camera pose embeddings, improving camera pose regression and consequently enhancing 3D reconstruction quality without any ground-truths of camera poses.
Our experiments demonstrate that the proposed approach achieves state-of-the-art performance in both standard GAN evaluation and 3D reconstruction metrics without extra information.
Our main contributions are summarized below:
\begin{itemize}
\item We present a simple yet effective camera pose representation method, implicit pose embedding, for training discriminators of 3D-aware GANs without ground-truth camera poses.
\item We train the discriminator of 3D-aware GANs by contrastive learning, which allows our model to learn 3D structures of scenes with ill-defined canonical poses due to heterogeneous geometric configurations.
\item Our framework achieves state-of-the-art performance on challenging benchmarks without any 3D related information and is validated via extensive experiments.
\end{itemize}

%% file: sections/_2_related_work.tex
% !TEX root = ../main.tex

\section{Related Work}
\label{sec:related}
We review existing approaches of GANs in 3D domain and discuss contrastive learning algorithms used in other generative tasks.

% 3D-aware GANs
\subsection{3D-aware GANs}
After the success of Generative Adversarial Network (GAN)~\cite{goodfellow2014generative,karras2019style,karras2020training, karras2020analyzing,Brock2018Biggan,zhang2019self,choi2020stargan} on generating high quality 2D images, several 3D-aware GANs~\cite{chan2022efficient,cai2022pix2nerf,niemeyer2021campari,chan2021pi,pan2021shading,or2022stylesdf,sun2022fenerf,deng2022gram,gu2021stylenerf,schwarz2020graf,skorokhodov2022epigraf} have been proposed to synthesize images based on 3D understanding instead of just image level understanding.
By plugging the ideas of volume rendering and neural implicit representation techniques~\cite{park2019deepsdf,niemeyer2020differentiable,mildenhall2021nerf} into networks, 3D-aware GANs gain the capability of synthesizing multiple views of a single 3D scene.
In addition, the networks are even trained to generate images with specific viewpoints using unorganized 2D image datasets---the same datasets used for 2D GANs.
However, the domains of such datasets are limited; most 3D-aware GANs have shown successful examples only in a few object classes, including human and animal faces, cars, and a few synthetic object categories. 
Unlike existing 3D-aware GANs, we extend the domain range of the 3D-aware GAN framework to the complex scenes.

\subsection{3D-aware GANs on complex scenes}
Some 3D-aware GANs have tackled generation task on the complex datasets which are composed of the images with diverse geometric configurations.
However, existing approaches mostly rely on the prior knowledge of scenes such as object classes, ground-truths camera poses, or depth supervision.
For example, full human body generation techniques~\cite{hong2022eva3d,grigorev2021stylepeople,zhang2022avatargen} demonstrate impressive results with high-fidelity geometries and motions but cannot be generalized to other domains because they rely on the pre-trained human body modeling such as SMPL~\cite{loper2015smpl}.
On another direction, several algorithms learn to generate 3D indoor environments~\cite{devries2021gsn,fraccaro2018gtm,bautista2022gaudi} and their generation processes are conditioned on camera pose information. 
These methods can synthesize new view synthesis of complex scenes in reasonable quality but require additional information such as ground truth camera poses or depth maps must be provided while training those networks.
DepthGAN~\cite{li2022depthgan} also generates 3D indoor environments, but it utilizes the estimated depth maps given by the pre-trained depth estimation model~\cite{yin2021learning} to obtain direct 3D information.
However, our algorithm does not rely on explicit geometric information such as depth maps or ground-truths camera poses, and it can be generalized to various domains.

\subsection{Contrastive learning}
Contrastive learning is a widely used self-supervised representation learning schemes~\cite{he2020momentum, wu2018unsupervised, chen2020simple,oord2018representation}.
Cntr-GAN~\cite{zhao2020image} adds contrastive learning to train GANs together with image augmentations, where it serves as a regularizer to improve the fidelity of generation.
Contrastive learning has also been used in image-to-image translation~\cite{park2020contrastive,han2021dual,ko2022self} and cross-modal translation~\cite{zhang2021cross} to enforce patch-wise correspondence and mutual information between image and text, respectively.
Also, ContraGAN~\cite{kang2020contragan} proposes a class-conditional contrastive learning objective to increase the correlations between images of the same class.
Unlike prior works, we are the first to adopt contrastive learning to 3D-aware GAN, employing it on the proposed implicit pose embeddings.

%% file: sections/_3_method.tex
% !TEX root = ../main.tex
\section{Preliminaries: EG3D}
\label{sec:preliminaries}

Our goal is to learn 3D-aware GANs on complex objects and scenes without prior knowledge or prediction models for camera poses of training examples.
Since our algorithm relies on a state-of-the-art model, EG3D~\cite{chan2022efficient}, we summarize the design of the generator $G(\cdot)$ and discriminator $D(\cdot)$ of EG3D. 

\begin{figure}[t]
    \centering
    \begin{subfigure}{.95\linewidth}
    \includegraphics[trim={0 0 0 0},clip,width=0.99\linewidth]{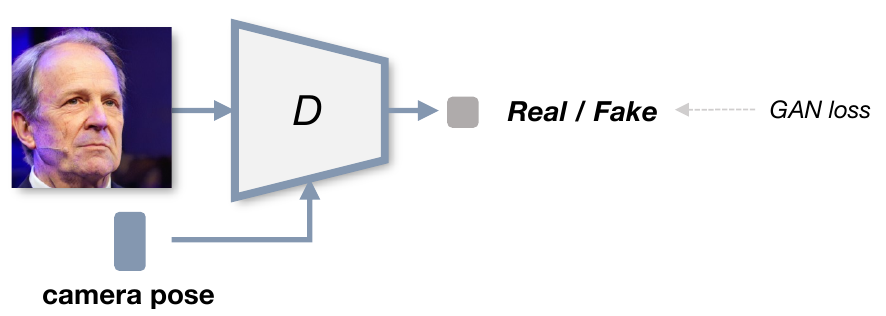}
    \caption{Pose-conditioned discriminator (EG3D)}
    \vspace{0.1cm}
    \label{fig:pose_conditioned_D}
    \end{subfigure}
    \begin{subfigure}{.95\linewidth}
    \includegraphics[trim={0 0 0 0},clip,width=0.99\linewidth]{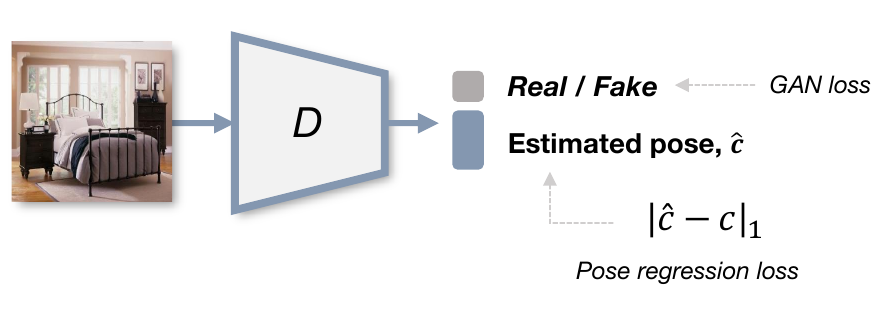}
    \caption{Camera pose regressing discriminator}
        \vspace{0.1cm}
    \label{fig:explicit_D}
    \end{subfigure}
    \begin{subfigure}{.95\linewidth}
    \includegraphics[trim={0 0 0 0},clip,width=0.99\linewidth]{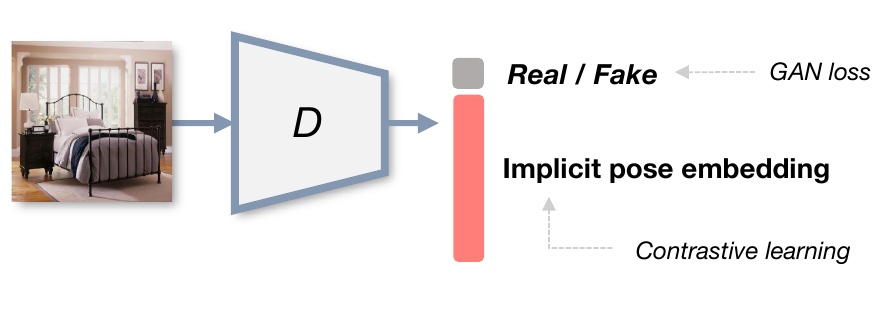}
    \vspace{-0.05cm}
    \caption{Implicit pose embedding discriminator}
        \vspace{0.1cm}
    \label{fig:implicit_D}
    \end{subfigure}
\caption{
Comparison of different discriminator architectures. 
The pose-conditioned discriminator in (a) utilizes camera pose information as input, where ground truth pose should be given for each training image. 
On the other hand, (b) and (c) does not use such extra information, but instead, they additionally learn camera pose estimator explicitly or implicitly on rendered images.
To this end, (b) uses direct pose regression loss with rendering camera pose ${c}$, while (c) employs contrastive learning to learn implicit pose embeddings.
Note that our PRNeRF and ContraNeRF employ (b) and (c) as their discriminators, respectively.
}
\label{fig:comparison}
\vspace{-0.2cm}
\end{figure}

\subsection{Generator}

Let $p_z$ and $p_{\xi}$ be the distributions of latent variable and camera pose, respectively.
Given $z \sim p_z$ and $c \sim p_\xi$, the generator produces a 3D feature based on a tri-plane structure.
The 3D feature is employed for rendering in the direction $c$, producing a low-resolution of 2D feature map and image.
Then, an image super-resolution module synthesizes a high-resolution image from a given 2D feature map and a low-resolution image.
In summary, the 3D-aware generator synthesizes a high-resolution image as
\begin{equation}
\label{eq:generator}
\begin{split}
G&: z, c \rightarrow I.
\end{split}
\end{equation}
The generator $G(\cdot)$ can produce an image with different viewpoints of the same object, \ie, using the identical $z$ but different $c$'s.

\subsection{Discriminator}
The discriminator in 3D-aware GANs encourages the paired generator to draw realistic images given camera poses.
To this end, EG3D~\cite{chan2022efficient} utilizes pose conditional discriminator as illustrated in Figure~\ref{fig:pose_conditioned_D}.
The discriminator takes both an image and a camera pose, and returns a logit as follows:
\begin{equation}
\label{eq:EG3D_discriminator}
\begin{split}
D&: I, c \rightarrow l,
\end{split}
\end{equation}
where $l \in \mathbb{R}$ is a logit for the standard GAN loss.
Note that, following the design of EG3D, the discriminator takes both low-resolution and high-resolution image as its inputs.
However, for the simplicity of the notations, we disregard low-resolution image inputs from the equations for EG3D and our algorithm, which include \eqref{eq:EG3D_discriminator}, ~\eqref{eq:explicit_discriminator}, and ~\eqref{eq:implicit_discriminator}.
Please refer to~\cite{chan2022efficient} for more detailed information.

\subsection{Discussion}
While EG3D~\cite{chan2022efficient} achieves competitive performance, it requires the ground-truth poses of training examples as inputs to the discriminator.
This limitation significantly reduces the applicability of EG3D because camera poses are defined relatively with respect to a certain viewpoint and consequently ill-defined except for a few object categories with common-sense central poses such as faces.
Other methods~\cite{niemeyer2021campari,chan2021pi,pan2021shading, sun2022fenerf, or2022stylesdf} trained without ground-truth camera poses typically yield incompetent performance and are evaluated only on less challenging datasets, \eg, human faces.
We propose a novel 3D-aware GAN algorithm that does not require camera pose labels but works well on images of natural scenes with heterogeneous geometric configurations.

\section{Camera Pose Regression in Discriminator}
\label{sec:camera}

This section describes our intermediate solution for training 3D-aware GAN models on top of EG3D without using camera pose labels of training data.

\subsection{Discriminator design}
\label{sub:discriminator_design}
To make the discriminator trainable without ground-truth camera poses, we first revise the original discriminator in EG3D~\cite{chan2022efficient} as illustrated in Figure~\ref{fig:explicit_D}.
Specifically, the new discriminator takes camera poses as its inputs no more while having additional output branch to predict the pose.
The formal definition of the discriminator operation is given by
\begin{equation}
\label{eq:explicit_discriminator}
\begin{split}
D&: I \rightarrow l, \hat{c},
\end{split}
\end{equation}
where $l \in \mathbb{R}$ is a logit for the standard GAN loss and $\hat{c} \in \mathbb{R}^2$ is an estimated pitch and yaw for camera pose of an input image $I$.
For implementation, we remove the pose conditional module in the discriminator of EG3D~\cite{chan2022efficient} and modify the dimension of the last fully connected layer.
Note that the generator has an identical architecture with EG3D~\cite{chan2022efficient}.

\subsection{Pose regression loss}
\label{sub:pose}
The estimated $\hat{c}$ is optimized by the pose regression loss, which encourages the generator to synthesize the images congruent to a given camera direction $c$, which is given by
\begin{equation}
\label{eq:pose_reconstruction}
\mathcal{L}_\text{pose}= \mathbb{E}_{z \sim p_z, c \sim p_{\xi}} \left \| \hat{c}-c \right \|,
\end{equation}
where $\left \| \cdot \right \|$ denotes $\ell_1$ or $\ell_2$ norm.
This loss is operated only on fake images because their true camera poses, denoted by $c$'s, are always available as rendering direction while we do not have ground-truths of real images.

\subsection{Overall objective}
\label{sub:overall_gan}
To enable the generator to learn the real data distribution, we use unsaturated adversarial losses with R1 regularization~\cite{mescheder2018training}.
On top of the standard GAN loss, the pose regression loss is employed to provide 3D awareness with the model as follows:
\begin{align}
\label{eq:pose_regression_loss}
\mathcal{L}(D,G) &= {\mathbb{E}}_{I\sim p_{\text{data}}}\left[f(-D(I) + \lambda \|\nabla D(I)\|^2)\right] \nonumber \\
&+{\mathbb{E}}_{z \sim p_z, c \sim p_{\xi}}\left[f(D(G(z, c))\right] \\
&+\lambda_\text{pose} \cdot \mathcal{L}_\text{pose}, \nonumber
\end{align}
where $f(u)=-\log(1+\exp(-u))$ and $p_{\text{data}}$ denotes the data distribution.

\section{Contrastive Learning in Discriminator}
\label{sec:contrastive}
We now discuss the formulation and optimization of our final model, ContraNeRF, for 3D-aware generative model via contrastive learning.

\subsection{Motivation}
\label{para:motivation}

Although the pose regression loss discussed in the previous section is effective in terms of generated image quality, it often suffers from a lack of fidelity in reconstructing the underlying 3D structures. 
We build a novel discriminator based on an implicit pose embedding in a high-dimensional space and train the network using a new loss based on pairwise relations of implicit camera poses in a mini-batch.
To be specific, we maximize the similarity between the implicit pose embeddings of the images with the same camera pose while minimizing the rest of the embedding pairs.
This strategy is helpful for encoding camera poses by estimating the underlying scene structures via rich and flexible relations between many implicit pose embedding pairs. 
Note that the pose regression loss is now inaccessible due to the use of implicit camera pose embedding but our approach still free from the ground-truth camera poses.

\subsection{Implicit pose embedding discriminator}
\label{sub:implicit}

The proposed discriminator has a similar architecture as the network given by~\eqref{eq:explicit_discriminator}.
The only difference is that, instead of extracting a two-dimensional explicit camera pose from the input image $I$, the new model estimates an high-dimensional implicit camera pose embedding as follows: 
\begin{equation}
\label{eq:implicit_discriminator}
\begin{split}
D&: I \rightarrow l, v, 
\end{split}
\end{equation}
where $l \in \mathbb{R}$ is a logit for the standard GAN loss and $v \in \mathbb{R}^m$ is an implicit pose embedding of the input image after $\ell_2$ normalization.
We set the dimensionality of $v$ sufficiently high, \ie $m \gg 2$, to make the implicit pose embedding vector more expressive than the typical camera pose representation based on yaw and pitch.
Figure~\ref{fig:comparison} illustrates our discriminator in comparison to other options.
The other parts of the discriminator are identical to EG3D~\cite{chan2022efficient}.

\subsection{Mutual information maximization}
\label{sub:mutual}
The idea of contrastive learning is to train a network to keep the representation of anchor images close to the representation of relevant positive images while pushing it away from those of many mismatched negative images.
Our goal is to make the synthesized images rendered on the same camera pose strongly associated with each other rather than the images generated by different poses as shown in Figure~\ref{fig:teaser}.
In this respect, we employ contrastive learning on the pose embedding in the discriminator, which aims to maximize the mutual information between synthesized images with the same camera pose.

\paragraph{Positive and negative examples}
Given an anchor image $I^a=G(z^a,c^a)$, a positive image $I^+$ and a negative image $I^-$ are defined as follows: 
\begin{align}
\label{eq:pos_neg_img}
I^+ &\in \mathcal{I}^+ = \{ I = G(z, c)  |  z \sim p_z, \ c = c^a \} \\
I^- &\in \mathcal{I}^- = \{ I = G(z, c)  |  z \sim p_z, \ c \sim p_\xi, \ c \neq c^a \}.
\end{align}
A positive image is an example that is rendered in the same direction but may be generated using a different latent vector from the anchor image.
In contrast, a negative image is the one rendered with a different camera pose from the anchor image.
Then, the implicit pose embedding of an anchor $v^a$, its positive embedding $v^+$ and negative embedding $v^-$ are given by
\begin{equation}
\label{eq:pos_neg_embeds}
\begin{split}
l^a, v^a & = D(I^a)
\\
l^+, v^+ & = D(I^+)
\\
l^-, v^- & = D(I^-),
\end{split}
\end{equation}
where $l^a$, $l^+$ and $l^-$ are logits for the standard GAN loss.

\paragraph{Contrastive loss}

We adopt the InfoNCE loss~\cite{oord2018representation} for contrastive learning of the implicit pose embedding.
Let $v_i^a$, $v_i^+$, and $v_{i,j}^a$ be camera pose embeddings in a mini-batch, where $v_i^+$ is a positive pair with the same camera pose of an anchor image as $v_i^a$, $i \in \left \{1, ..., N\right \}$, while $v_{i,j}^-$, $j \in \left \{1, ..., S\right \}$, is a negative pair with a different pose from $v_i^a$.
We denote a collection of the negative examples for each anchor by $\boldsymbol{v}_i^-$, \ie, $v_{i,j}^- \in \boldsymbol{v}^-$.
Given $v_i^a$, $v_i^+$, and $\boldsymbol{v}_i^-$, we obtain the following contrastive loss term:
\begin{align}
\label{eq:info_nce}
    &\ell\left(v_i^a, v_i^{+}, \boldsymbol v_i^{-}\right)=  \\
    &-\log \left( \frac{\exp \left(d({v_i^a},v_i^{+}) / \tau\right)}{\exp \left(d(v_i^a, v_i^{+}) / \tau\right)+\sum_{j=1}^{S} \exp \left( d(v_i^a, v_{i,j}^{-}) / \tau\right)} \right), \nonumber
\end{align}
where $d(u, v)= {u}^{\top} {v} /\|u\|\|{v}\|$ denotes the cosine similarity between ${u}$ and ${v}$.
This loss enforces the synthesized image to be similar as the rendered images in the same camera viewpoint but dissimilar to those in other camera directions.
To sum up, the overall contrastive loss is given by
\begin{equation}
\label{eq:posence}
\mathcal{L}_\text{InfoNCE} = {\mathbb{E}}_{z \sim p_z, c \sim p_{\xi}}\ell(v^a, v^+, \boldsymbol{v}^-),
\end{equation}
where $v^+$ and $\boldsymbol{v}^-$ are defined for each anchor, $v^a$.

\subsection{Overall Objective}
\label{sub:overall_pr}
The final objective function of our algorithm is given by replacing the pose regression term in~\eqref{eq:pose_regression_loss} by the InfoNCE loss term proposed for contrastive learning, as follows:
\begin{align}
\label{eq:final_loss}
\mathcal{L}(D,G) &= {\mathbb{E}}_{I\sim p_{\text{data}}}\left[f(-D(I) + \lambda \|\nabla D(I)\|^2)\right] \nonumber \\
&~~~~+{\mathbb{E}}_{\z \sim p_z, c \sim p_{\xi}}\left[f(D(G(z, c))\right] \\
&~~~~+\lambda_\text{pose} \cdot \mathcal{L}_\text{InfoNCE}, \nonumber
\end{align}
where the first term is active for real images, updating only the discriminator, while the remaining terms optimize both the generator and the discriminator with fake images.

%% file: sections/_4_experiment.tex
% !TEX root = ../main.tex
\section{Experiments}
\label{sec:experiments}

This section describes our benchmarks with complex geometric structures and reports the performance of our methods compared to previous ones quantitatively and qualitatively.
We referred to our two models, one with camera pose regression and the other with contrastive learning, as PRNeRF and ContraNeRF, respectively.

\subsection{Datasets and Settings}
We report results on four different image datasets, LSUN Bedroom~\cite{yu2015lsun}, LSUN Church~\cite{yu2015lsun}, AFHQ (Animal Faces-HQ)~\cite{choi2020stargan}, CUB~\cite{wah2011caltech}.
These datasets are challenging for 3D-aware GANs, where the canonical pose is hard to define on the LSUN datasets, and AFHQ and CUB datasets contain complex and diverse geometric shapes.
For AFHQ, we compute low-resolution features and images at a resolution of $32^2$ with a total of 96 depth samples per ray. 
The final images are generated at a resolution of $256^2$. 
The resolution of feature maps and final images for the other datasets is set to $32^2$ and $128^2$, respectively.

\subsection{Results}
Several ablations and analyses are performed to justify our contributions and proposed modules. 
For image synthesis evaluation, we report Fr\'echet inception distance (FID)~\cite{heusel2017gans} and Precision \& Recall, which measure the fidelity and diversity of generated samples.
For the evaluation of 3D reconstruction quality, we visualize rendered depth images along with their Depth FID, which measures FID between the estimated depth maps of training images given by a  depth estimation model~\cite{yin2021learning} and the rendered depth maps from the generated images.
We also provide the quality of depth in rendered images using three subjective levels: \textcolor{red}{Bad}, \textcolor{Green}{Fair}, and \textcolor{blue}{Good}.

%%%%%%%%%%%%%%%%%%%%%%%%%%%%%%%%%%%%%
%                             1) LSUN Bedroom 
%%%%%%%%%%%%%%%%%%%%%%%%%%%%%%%%%%%%%%
\subsubsection{LSUN Bedroom}

We compare PRNeRF and ContraNeRF with GRAF~\cite{schwarz2020graf}, GIRAFFE~\cite{niemeyer2021giraffe}, $\pi$-GAN~\cite{chan2021pi}, and DepthGAN~\cite{li2022depthgan} on the LSUN bedroom dataset. 
Figure~\ref{fig:qualitative_all}\textcolor{red}{a} illustrates the generated images and their depth maps from three different viewpoints.
Most algorithms including $\pi$-GAN~\cite{niemeyer2021giraffe}, GIRAFFE~\cite{schwarz2020graf}, and PRNeRF produce unrealistic depth maps, where their generated images are almost identical and have planer depth maps from all viewpoints.
In contrast, ContraNeRF generates RGB images and depth maps that reflect true 3D scene structures faithfully. 
Table~\ref{tab:lsun_bedroom} presents overall quantitative results, where ContraNeRF outperforms other algorithms in terms of Depth FID with considerable margins. 
It indicates that the synthesized 3D scenes given by ContraNeRF reflect true geometries effectively.
Within our methods, although PRNeRF outperforms ContraNeRF in terms of 2D image synthesis metrics, it struggles with learning 3D structures accurately.

\begin{table}[t]
\centering
\caption{Quantitative comparison on the LSUN Bedroom dataset.
Our models outperform existing methods by significant margins in all image quality metrics, where ContraNeRF successfully learns 3D geometry information.}
\vspace{-2mm}
\label{tab:lsun_bedroom}
\scalebox{0.85}{
\setlength\tabcolsep{2.5pt}
\hspace{-3mm}
\begin{tabular}{ccccccc}
\toprule
Method	&FID $\downarrow$ & Precision $\uparrow$ & Recall $\uparrow$ & Depth FID$\downarrow$ & 3D\\ 
\hline 
\hline
GRAF~\cite{schwarz2020graf}					&70.71 	& 0.42 	& 0.00 	&97.41	&\text{\textcolor{red}{Bad}} \\
$\pi$-GAN~\cite{chan2021pi}		  			&56.32	& 0.44 	& 0.11  	&124.10	&\text{\textcolor{red}{Bad}} \\
GIRAFFE~\cite{niemeyer2021giraffe}			&42.78 	& 0.55	& 0.02 	&145.63 	&\text{\textcolor{red}{Bad}} \\
\hdashline
PRNeRF (ours) 							&14.97 	& 0.55 	& 0.19 	&110.42 	&\text{\textcolor{red}{Bad}}\\
ContraNeRF (ours)               					&15.31 	& 0.54 	& 0.15 	& 49.30  	& \text{\textcolor{blue}{Good}}\\
\bottomrule
\end{tabular}}
\vspace{-2mm}
\end{table}

%%%%%%%%%%%%%%%%%%
% 2) LSUN Church
%%%%%%%%%%%%%%%%%%
\subsubsection{LSUN Church}
Our models are compared with GRAF~\cite{schwarz2020graf}, GIRAFFE~\cite{niemeyer2021giraffe}, $\pi$-GAN~\cite{chan2021pi}, and GIRAFFE-HD~\cite{xue2022giraffe} on the LSUN church dataset.
Again, our models outperform previous models in all metrics, as shown in Table~\ref{tab:lsun_church}.
Figure~\ref{fig:qualitative_all}\textcolor{red}{b} illustrates output examples from our models, where only ContraNeRF produces reasonable 3D scene structures and images.
Similar to LSUN Bedroom, although PRNeRF shows a better 2D image synthesis quality within our methods, it fails to capture realistic 3D information in the scene.

\begin{table}[t]
\centering
\caption{Quantitative comparison on the LSUN Church dataset. 
Our models outperform all other existing methods by significant margins in all image quality metrics, where ContraNeRF achieves the best performance.
The dagger ($\dagger$) denotes that the scores are taken from GIRAFFE-HD~\cite{xue2022giraffe}.}
\vspace{-2mm}
\label{tab:lsun_church}
\scalebox{0.85}{
\setlength\tabcolsep{7pt}
\hspace{-2mm}
\begin{tabular}{cccccccc}
\toprule
Method & FID $\downarrow$ & Precision $\uparrow$ & Recall $\uparrow$ & 3D \\ \hline \hline
GRAF~\cite{schwarz2020graf}			                        	& 91.11 	   	& 0.53	  &0.00  	&\text{\textcolor{red}{Bad}}\\
$\pi$-GAN~\cite{chan2021pi}			    			& 56.80 		& 0.49        & 0.18 	&\text{\textcolor{red}{Bad}} \\
GIRAFFE~\cite{niemeyer2021giraffe}		                 & 38.36 	  	& 0.51        & 0.02 	&\text{\textcolor{red}{Bad}}\\ %large basline
GIRAFFE-HD~\cite{xue2022giraffe}$\dagger$			& 10.28 		& --         & -- 	& --\\
\hdashline
PRNeRF (ours)  								& \ \ 5.48  		& 0.58     & 0.40	&\textcolor{red}{Bad} \\  %2dim
ContraNeRF (ours)               						& \ \ 5.94  		& 0.61     & 0.36  	&\textcolor{blue}{Good}\\
\bottomrule
\end{tabular}}
\end{table}

%%%%%%%%%%%%%%%%%%
%                  3) AFHQ
%%%%%%%%%%%%%%%%%%
\subsubsection{AFHQ: Cats, Dogs, Wildlifes}
AFHQ includes three categories for the animal face of \textit{cats}, \textit{dogs}, and \textit{wildlife}, and we merge the three subsets and construct a single dataset having more diverse object shapes.
Figure~\ref{fig:qualitative_all}\textcolor{red}{c} presents the images of animal faces generated from different viewpoints, and Table~\ref{tab:afhq} summarizes the experimental results on AFHQ.
Both PRNeRF and ContraNeRF exhibit reasonable 3D-aware image synthesis performance, but ContraNeRF works slightly better in all evaluation metrics. 
PRNeRF does not produce planar depth maps, unlike the LSUN datasets, because the geometric structures of the AFHQ dataset are more straightforward.
\begin{table}[t]
\centering
\caption{
Quantitative comparison on the unified AFHQ dataset, including the Cats, Dogs, and Wilds categories with $256^2$ resolution.
PRNeRF and ContraNeRF synthesize high-fidelity images capturing accurate 3D geometry, where ContraNeRF shows the best performance.
The dagger ($\dagger$) denotes that the scores are taken from StyleNeRF~\cite{gu2021stylenerf}.}
\vspace{-2mm}
\label{tab:afhq}
\scalebox{0.85}{
\setlength\tabcolsep{7pt}
\hspace{-2mm}
\begin{tabular}{ccccc}
\toprule
Method & FID $\downarrow$ & Precision $\uparrow$ & Recall $\uparrow$ & 3D\\
\hline  
\hline
GRAF~\cite{schwarz2020graf} 					&107.00 \ \		&0.35	&0.00		&\textcolor{red}{Bad}\\
$\pi$-GAN~\cite{chan2021pi}					&48.43			&0.41	&0.12		&\textcolor{red}{Bad}\\
GIRAFFE~\cite{niemeyer2021giraffe}	 		&31.31			&0.51	&0.04	&\textcolor{Green}{Fair}\\
GIRAFFE-HD~\cite{xue2022giraffe}				&14.21			&0.55	&0.10	&--\\
StyleNeRF~\cite{gu2021stylenerf}$^\dagger$		&14 \ \ \ \ \	 		&--		&--		&\textcolor{blue}{Good} \\
\hdashline
PRNeRF (ours)  							&\ \ 9.14 			&0.54 	&0.19 	&\textcolor{blue}{Good}\\
ContraNeRF (ours)            					&\ \ 9.03			&0.55	&0.21	&\textcolor{blue}{Good} \\
\bottomrule
\end{tabular}}
\end{table}
%
%
%
%

%%%%%%%%%%%%%%%%%%
%                 4) CUB
%%%%%%%%%%%%%%%%%%
\subsubsection{CUB}

\begin{figure*}[t]
     \centering
     \begin{flushleft}
    \small \ \ \ \ \ \ \ \ \ \ \ \ \ \ $-40^{\circ}$ \ \ \ \ \ \ \ \  $0^{\circ}$  \ \ \ \ \ \ \ \ \  $40^{\circ}$ 
              \ \ \ \ \ \ \ \ \ \ \ \ \ $-40^{\circ}$ \ \ \ \ \ \ \ \  $0^{\circ}$  \ \ \ \ \ \ \ \ \ $40^{\circ}$
              \ \ \ \ \ \ \ \ \ \ \ \ \ \ $-30^{\circ}$ \ \ \ \ \ \ \ \  $0^{\circ}$  \ \ \ \ \ \ \ \ \ $30^{\circ}$
              \ \ \ \ \ \ \ \ \ \ \ \ \ $-40^{\circ}$ \ \ \ \ \ \ \ \  $0^{\circ}$  \ \ \ \ \ \ \ \ \ $40^{\circ}$           
    \vspace{-3mm} 
     \end{flushleft}
     \vfill \hfill
     \begin{minipage}{0.03\linewidth}
     \rotatebox[origin=l]{90}{
     \small   ContraNeRF (ours) \ \ \ \ \  \ PRNeRF (ours) \ \ \ \ \ \ \ \ \ GIRAFFE~\cite{niemeyer2021giraffe}  \ \ \ \ \ \ \ \  \ \ \ \ \ \ \ \cite{chan2021pi} \cite{xue2022giraffe} \ \ \
     }
     \end{minipage}
     \hspace{-3mm}
	\begin{minipage}{0.975\linewidth}
     \begin{subfigure}[b]{0.23\linewidth}
         \centering
         \includegraphics[width=0.99\linewidth]{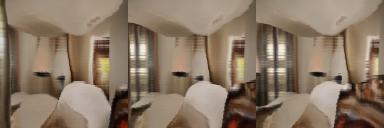}
          \vspace{1mm}
         \includegraphics[width=0.99\linewidth]{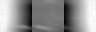}
      	 \hfill
         \includegraphics[width=0.99\linewidth]{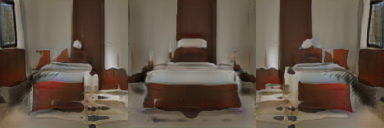}
          \vspace{1mm}
         \includegraphics[width=0.99\linewidth]{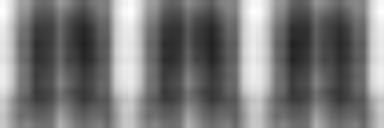}
         \hfill
         \includegraphics[width=0.99\linewidth]{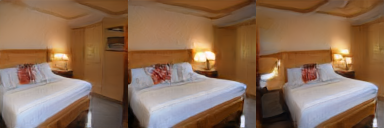}
          \vspace{1mm}
         \includegraphics[width=0.99\linewidth]{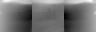}
         \hfill
         \includegraphics[width=0.99\linewidth]{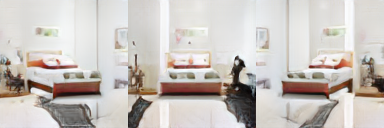}
         \includegraphics[width=0.99\linewidth]{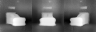}
         \caption{LSUN Bedroom}
     \end{subfigure}
     \hspace{1mm}
          \begin{subfigure}[b]{0.23\linewidth}
         \centering
         \includegraphics[width=0.99\linewidth]{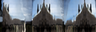}
         \vspace{1mm}
         \includegraphics[width=0.99\linewidth]{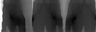}
         \hfill
         \includegraphics[width=0.99\linewidth]{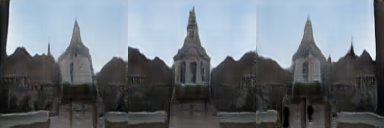}
          \vspace{1mm}
         \includegraphics[width=0.99\linewidth]{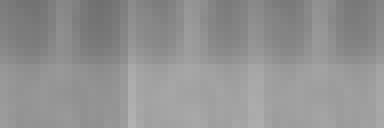}
         \hfill
         \includegraphics[width=0.99\linewidth]{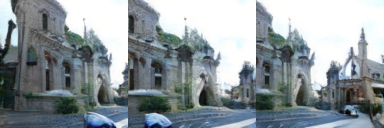}
          \vspace{1mm}
         \includegraphics[width=0.99\linewidth]{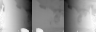}
         \hfill
         \includegraphics[width=0.99\linewidth]{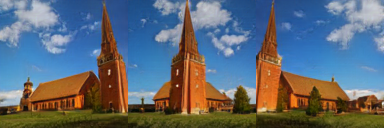}
         \includegraphics[width=0.99\linewidth]{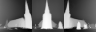}
         \caption{LSUN Church}
       	\end{subfigure}
	\hspace{1mm}
        	          \begin{subfigure}[b]{0.23\linewidth}
         \centering
         \includegraphics[width=0.99\linewidth]{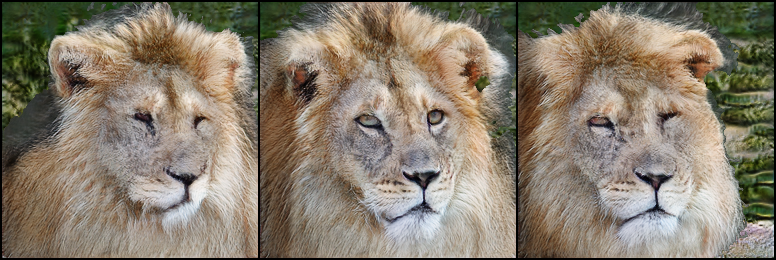}
         \vspace{1mm}
         \includegraphics[width=0.99\linewidth]{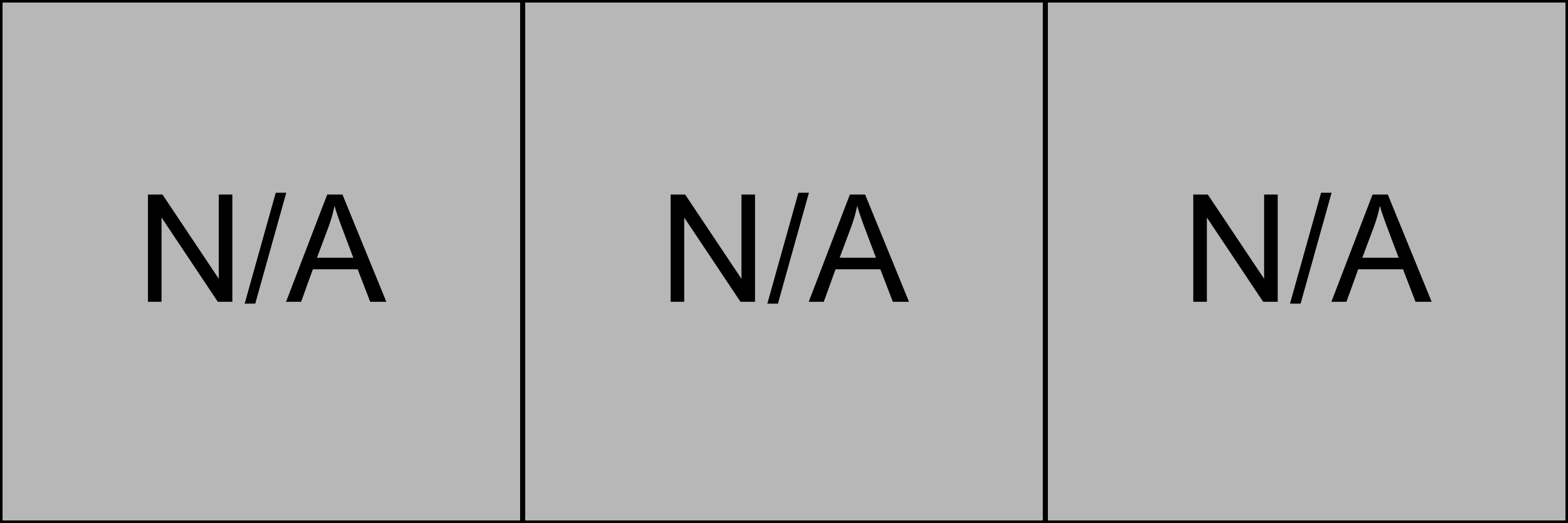}
         \hfill
         \includegraphics[width=0.99\linewidth]{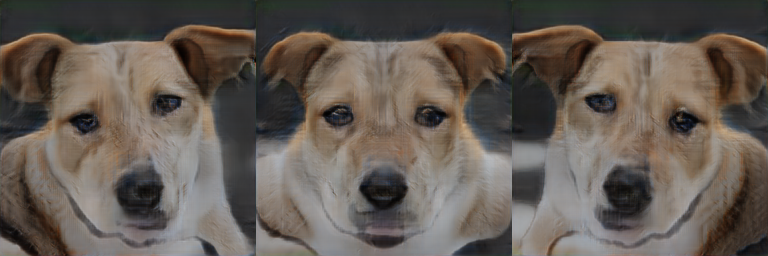}
          \vspace{1mm}
         \includegraphics[width=0.99\linewidth]{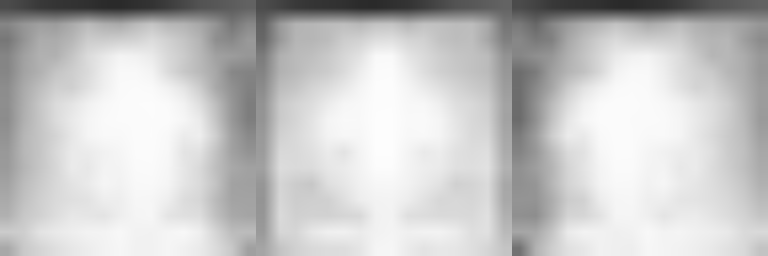}
         \hfill
         \includegraphics[width=0.99\linewidth]{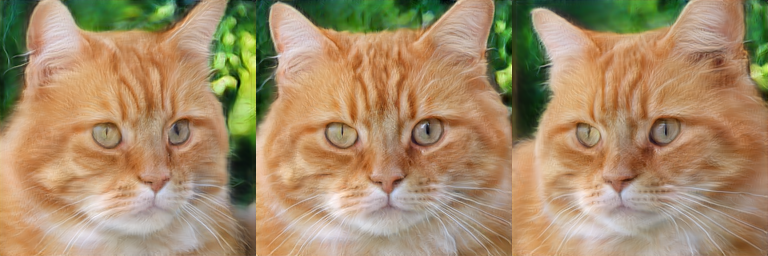}
         \vspace{1mm}
         \includegraphics[width=0.99\linewidth]{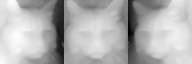}
         \hfill
         \includegraphics[width=0.99\linewidth]{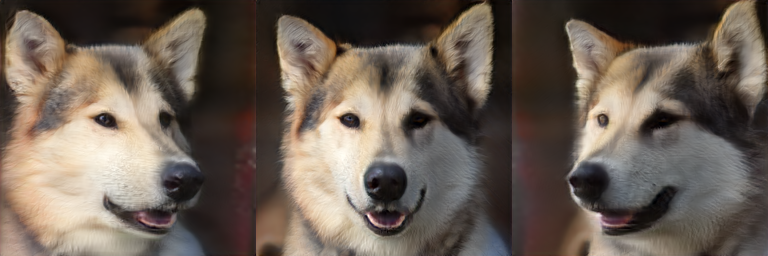}
         \includegraphics[width=0.99\linewidth]{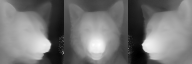}
         \caption{AFHQ (All categories)}
       	\end{subfigure}
	\hspace{1mm}
	 \begin{subfigure}[b]{0.23\linewidth}
         \centering
        \includegraphics[width=0.99\linewidth]{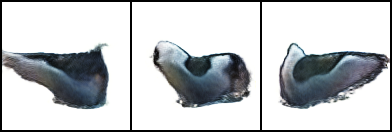}
         \vspace{1mm}
         \includegraphics[width=0.99\linewidth]{images/all_comparison/na.png}
         \hfill
         \includegraphics[width=0.99\linewidth]{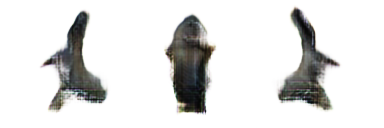}
          \vspace{1mm}
         \includegraphics[width=0.99\linewidth]{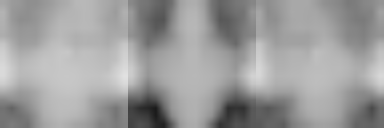}
         \hfill
         \includegraphics[width=0.99\linewidth]{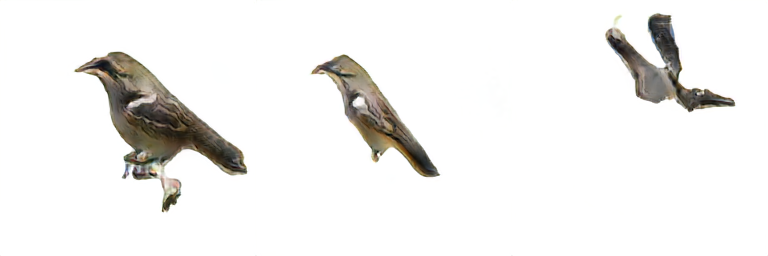}
          \vspace{1mm}
         \includegraphics[width=0.99\linewidth]{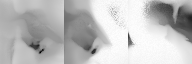}
         \hfill
         \includegraphics[width=0.99\linewidth]{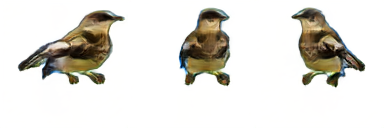}
         \includegraphics[width=0.99\linewidth]{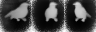}
         \caption{CUB}
       	\end{subfigure}
	\end{minipage}
	\hspace{3mm}
	\vspace{-1mm}
        \caption{
	Comparing samples of ContraNeRF, PRNeRF, and modern 3D-aware GANs.
	We visualized the RGB image and depth map by rotating the rendering pose horizontally.
	For the first row, the first two results are from $\pi$-GAN~\cite{chan2021pi}, and the rest is from GIRAFFE-HD~\cite{xue2022giraffe}.
	ContraNeRF produces high-fidelity images with accurate depth maps in all domains since its implicit pose embedding can capture complex geometries.
   	However, others methods, including PRNeRF, usually produce planar depth maps with unrealistic 3D structure. 
	For additional results, please refer to our supplementary materials.}
        \label{fig:qualitative_all}
        \vspace{-2mm}
\end{figure*}

We evaluate our framework in a more challenging dataset, CUB, which consists of images with large object pose variations.
Table~\ref{tab:cub} presents that ContraNeRF still outperforms others by large margins, offering the lowest FID values.
This is because our pose embedding captures rich geometric information of scenes from diverse pairwise relations of camera pose embeddings. 
As illustrated in Figure~\ref{fig:qualitative_all}\textcolor{red}{d}, the quality of the depth maps estimated by ContraNeRF looks impressive, and we notice that all the compared algorithms mostly fail to reconstruct 3D structures accurately. 

\begin{table}[t]
\centering
\caption{Quantitative comparison on the CUB dataset, having large object pose variations.
ContraNeRF significantly outperforms existing methods in all image quality metrics.}
\vspace{-2mm}
\label{tab:cub}
\scalebox{0.85}{
\setlength\tabcolsep{8pt}
\hspace{-2mm}
\begin{tabular}{cccccccc}
\toprule
Method & FID $\downarrow$ & Precision $\uparrow$ & Recall $\uparrow$ & 3D
\\ 
\hline 
\hline
GRAF~\cite{schwarz2020graf} 								&46.31 	&0.67	&0.09	  &\textcolor{red}{Bad}\\
GIRAFFE~\cite{niemeyer2021giraffe}                        				&49.34 	&0.68	&0.04 		&\textcolor{red}{Bad}\\
$\pi$-GAN~\cite{chan2021pi}								&48.82	&0.64	&0.10		&\textcolor{red}{Bad}\\
GIRAFFE-HD~\cite{xue2022giraffe}                       				& 24.31	&0.67	&0.17 	& -- \\
\hdashline
PRNeRF (ours)  										&48.34 	  &0.65	&0.08	&\textcolor{red}{Bad}\\\
ContraNeRF (ours)               								&17.65	&0.71	&0.24	&\textcolor{blue}{Good}\\
\bottomrule
\end{tabular}}
\vspace{-4mm}
\end{table}

\subsection{Analysis}
%%%%%%%%%%%%%%%%%%%%%%%
%             Comparison with EG3D
%%%%%%%%%%%%%%%%%%%%%%%%%
%\vspace{-2mm}
\paragraph{Combination of $\mathcal{L}_\text{Pose}$ and $\mathcal{L}_\text{InfoNCE}$}
We evaluate the ensemble trained with the pose regression loss, $\mathcal{L}_\text{Pose}$, and the contrastive loss, $\mathcal{L}_\text{InfoNCE}$, on the FFHQ dataset~\cite{karras2019style}.
Table~\ref{tab:result_ffhq} presents that the combination of the two losses achieves the best performance on FFHQ, except EG3D\footnote{EG3D exploits the ground-truth camera poses of the training set, which makes the direct comparison with EG3D unfair. For reference, EG3D can not be evaluated by other datasets used in this paper.}.
Unlike other datasets, PRNeRF outperforms ContraNeRF in FFHQ, probably because the pose regression of PRNeRF is more straightforward in this dataset, having homogeneous geometry.
However, ContraNeRF or its ensemble version always shows the best performance including the FFHQ dataset.

\begin{table}[t]
        	\centering
    	\caption{Experiments on the FFHQ dataset with 256 resolution.
	The dagger ($\dagger$) denotes that the scores are taken from StyleNeRF~\cite{gu2021stylenerf} and EpiGRAF~\cite{skorokhodov2022epigraf}.}
	\vspace{-2mm}
	\label{tab:result_ffhq}
	\scalebox{0.85}{
   	\setlength\tabcolsep{2.5pt} 
	\hspace{-3mm}
    	\begin{tabular}{c|c|cc|ccc}
    	\toprule
    	Method & GT pose &  $\mathcal{L}_\text{Pose}$	& $\mathcal{L}_\text{InfoNCE}$ &FID$\downarrow$ &Precision$\uparrow$ &Recall$\uparrow$ \\
	\hline
	\hline
	EG3D~\cite{chan2022efficient} 			& \checkmark	&			&			&4.92		&0.554		&0.435\\
	\hdashline
	StyleNeRF~\cite{gu2021stylenerf}$^\dagger$ & 			&			&							&8 \ \ \ \ \ 		&-			&-\\
	EpiGRAF~\cite{skorokhodov2022epigraf}$^\dagger$	   & 			&			&							&9.71		&-			&- \\
	PRNeRF 			&			& \checkmark 	&			&5.94		&0.548		&0.415\\
	ContraNeRF 		&			& 			& \checkmark 	&6.85		&\textbf{0.552}	&0.405\\
	PR-ContraNeRF	& 			& \checkmark 	& \checkmark 	&\textbf{5.73}	&0.551		&\textbf{0.421}\\
	\bottomrule
    	\end{tabular}
    	}
	\vspace{-2mm}
\end{table}

%%%%%%%%%%%%%%%%%%%%%%%
%                 High resolution
%%%%%%%%%%%%%%%%%%%%%%%%%
\vspace{-2mm}
\paragraph{High resolution image synthesis}
To verify that our algorithm performs well on higher-resolution images, we test our algorithms on the AFHQ dataset with the resolution of  $512^2$.
Table~\ref{tab:resol_512} presents that our methods still significantly outperform the previous one on the $512^2$ resolution, similar to the $256^2$ resolution setting in Table~\ref{tab:afhq}. 

%Table: res 512
\begin{table}[t]
        	\centering
    	\caption{Experiments on the AFHQ dataset with $512^2$ resolution. ContraNeRF and PRNeRF still outperform the existing method in high resolution setting, where ContraNeRF achieves the best performance.}
	\vspace{-2mm}
    	\scalebox{0.85}{
   	\setlength\tabcolsep{13pt} 
	\hspace{-2mm}
    	\begin{tabular}{c|cccc}
    	\toprule
    	Method 			& FID$\downarrow$. &Precision$\uparrow$ &Recall$\uparrow$\\
	\hline
	\hline
	GIRAFFE-HD~\cite{xue2022giraffe} 		&13.42			&0.61		&0.23\\
	\hdashline
	PRNeRF							&\ \ 8.21			&0.61		&0.31\\
	ContraNeRF						&\ \ \textbf{8.02}		&\textbf{0.63}	&\textbf{0.32}\\
	\bottomrule
    	\end{tabular}
    	}
\label{tab:resol_512}
\end{table}

%%%%%%%%%%%%%%%%%%%%%%%
%             dimension of pose embedding
%%%%%%%%%%%%%%%%%%%%%%%%%
\vspace{-2mm}
\paragraph{Dimensionality of pose embedding}
\begin{figure}
     \centering
     \begin{subfigure}[b]{0.49\linewidth}
         \centering
         \includegraphics[width=0.50\linewidth]{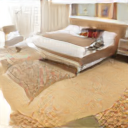}%
         \includegraphics[width=0.50\linewidth]{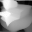}
         \caption{2 dimension: \textcolor{Green}{Fair}}
         \label{fig:qual_ablation_2dim}
     \end{subfigure}
     \begin{subfigure}[b]{0.49\linewidth}
         \centering
         \includegraphics[width=0.50\linewidth]{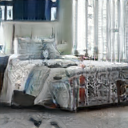}%
         \includegraphics[width=0.50\linewidth]{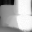}
         \caption{12 dimension: \textcolor{Green}{Fair}}
          \label{fig:qual_ablation_12dim}
     \end{subfigure}
     \begin{subfigure}[b]{0.49\linewidth}
         \centering
         \includegraphics[width=0.50\linewidth]{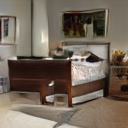}%
         \includegraphics[width=0.50\linewidth]{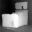}
         \caption{24 dimension: \textcolor{blue}{Good}}
         \label{fig:qual_ablation_24dim}
     \end{subfigure}
     \begin{subfigure}[b]{0.49\linewidth}
         \centering
         \includegraphics[width=0.50\linewidth]{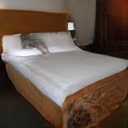}%
         \includegraphics[width=0.50\linewidth]{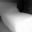}
         \caption{48 dimension: \textcolor{blue}{Good}}
         \label{fig:qual_ablation_48dim}
     \end{subfigure}
     \vspace{-0.1cm}
        \caption{Effects of the pose embedding dimension on the quality of rendered image on the LSUN Bedroom dataset.
        Our framework successfully captures underlying 3D structures with a sufficient number of embedding dimensions.
        }
        \label{fig:ablation_dim}
\end{figure}

We analyze the impacts of the dimensionality of pose embedding on 3D reconstruction quality on the LSUN Bedroom dataset, and visualize its ablative results in Figure~\ref{fig:ablation_dim}.
Our framework successfully captures 3D structures if it has a sufficient number of embedding dimension $m \geq 24$.
Even with low-dimensional embedding vectors, we still obtain decent quality of reconstructed images and depth maps with minor blurs.

%%%%%%%%%%%%%%%%%%%%%%%
%                same pose, differ z
%%%%%%%%%%%%%%%%%%%%%%%%%
\vspace{-2mm}
\paragraph{Handling dataset with diverse camera poses} 
\begin{figure}
     \centering
     \begin{subfigure}[b]{0.49\linewidth}
         \centering
         \includegraphics[width=\linewidth]{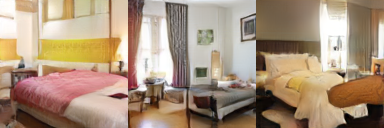}
         \caption{LSUN Bedroom}
         \vspace{1mm}
     \end{subfigure}
     \hfill
     \begin{subfigure}[b]{0.49\linewidth}
         \centering
         \includegraphics[width=\linewidth]{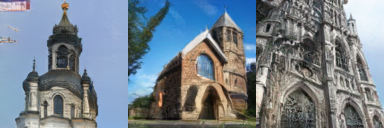}
         \caption{LSUN Church}
         \vspace{1mm}
     \end{subfigure}
     \hfill
     \begin{subfigure}[b]{0.49\linewidth}
         \centering
         \includegraphics[width=\linewidth]{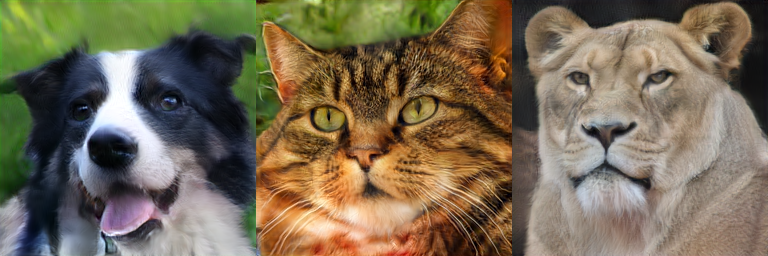}
        \caption{AFHQ (All categories)}   
     \end{subfigure}
          \hfill
     \begin{subfigure}[b]{0.49\linewidth}
         \centering
         \includegraphics[width=\linewidth]{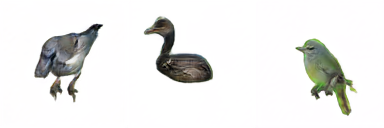}
        \caption{CUB}   
     \end{subfigure}
     \vspace{-2mm}
        \caption{Qualitative results of rendering for the same pose with different latent vector $z$ on ContraNeRF.
	      ContraNeRF synthesizes diverse scenes with different geometry in the LSUN Bedroom and LSUN Church datasets while it produces images with identical geometry in the AFHQ dataset.
	        }
        \label{fig:same_pose}
\end{figure}

Figure~\ref{fig:same_pose} illustrates images rendered by ContraNeRF, from the same camera pose but with different content latent vector $z$.
The generated images on the AFHQ dataset have almost identical viewpoints, indicating that our contrastive learning works as previous ones for these simple cases.
On the other hand, since the LSUN Bedroom, LSUN Church, and CUB have various scene geometries and object shapes, there is no canonical center pose applicable to all images, and the generated images do not have the same viewpoints perceptually.
However, ContraNeRF successfully reconstructs the underlying 3D structure of the scenes as presented earlier, which shows the strength and potential of ContraNeRF for naturally handling datasets with images captured from heterogeneous viewpoints.

%%%%%%%%%%%%%%%%%%%%%%%
%                  Failure case
%%%%%%%%%%%%%%%%%%%%%%%%%
\vspace{-2mm}
\paragraph{Failure cases}

\begin{figure}[t]
     \centering
	\begin{subfigure}[b]{0.48\linewidth}
	\centering
	\includegraphics[width=\linewidth]{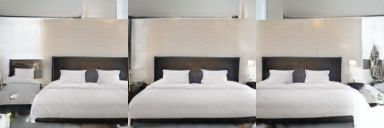}
	\includegraphics[width=\linewidth]{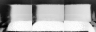}
 	\end{subfigure}
	\hspace{1mm}
	\begin{subfigure}[b]{0.48\linewidth}
	\centering
         \includegraphics[width=\linewidth]{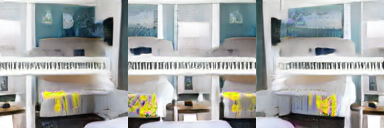}
         \includegraphics[width=\linewidth]{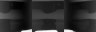}
	\end{subfigure}
	\vspace{-2mm}
        \caption{
    Example of failure cases. ContraNeRF sometimes produces images with unrealistic geometries such as planar scenes. 
    The first case (left) is the example that our algorithm generate translated images by varying camera poses, and the second one (right) illustrates the results with almost uniform depth maps. 
        }\label{fig:failure_case}
\end{figure}

Although ContraNeRF produces high-fidelity volumetric scenes in most cases, we observe some failure cases on the LSUN bedroom dataset. 
Figure~\ref{fig:failure_case} illustrates failure cases in which ContraNeRF produces planar scenes.
We presume outlier training samples, such as watermarked images or images captured from out-of-distribution camera poses, may result in degenerate outputs.

%% file: sections/_5_conclusion.tex
% !TEX root = ../main.tex
\section{Conclusion}
\label{sec:conclusion}
By extending 3D-aware GANs to handle more diverse domains of objects and scenes, the proposed models improve their usability and expands the possible applications from face synthesis to 3D world modeling.
To this end, we first show that a pose regression-based framework can be used to effectively remove the camera pose dependency in 3D-aware GAN training.
We then propose a contrastive learning-based framework that uses implicit pose embeddings at higher dimensions for rich descriptions of pose information in natural scenes with diverse and complex geometries.
The effectiveness of implicit pose embedding and contrast learning frameworks has been experimentally demonstrated through evaluation on multiple benchmark datasets.

%% file: sections/_supple.tex
% !TEX root = ../main.tex
\onecolumn
\setcounter{section}{0}
\setcounter{table}{0}
\setcounter{figure}{0}
\renewcommand\thesection{\Alph{section}}
\renewcommand\thetable{\Alph{table}}
\renewcommand\thefigure{\Alph{figure}}

\section{Ensemble of $\mathcal{L}_{\text{Pose}}$ and $\mathcal{L}_{\text{InfoNCE}}$}

We further evaluate the combination of pose regression loss, $\mathcal{L}_{\text{Pose}}$, and contrastive loss, $\mathcal{L}_{\text{InfoNCE}}$, on the AFHQ and LSUN Bedroom datasets, with comparative results presented in Table~\ref{tab:ablation_combination_afhq}.
For the AFHQ dataset, the combined approach enhances 2D image synthesis metrics without compromising 3D reconstruction quality, though the differences are minor. 
Conversely, on the LSUN Bedroom dataset, the ensemble leads to a slight decline in 3D reconstruction quality, partly due to the pose regression loss being ill-suited for representing complex scene structures.

\vspace{2mm}
\begin{table}[h]
\centering
\caption{Ablative results of pose regression loss and contrastive loss on the AFHQ and LSUN Bedroom dataset.}
\label{tab:ablation_combination_afhq}
\scalebox{0.9}{
\setlength\tabcolsep{7pt}
\begin{tabular}{c|cc|cccc|cccc}
\toprule
\multirow{2}{*}{Algorithms} &\multirow{2}{*}{$\mathcal{L}_\text{Pose}$} &\multirow{2}{*}{$\mathcal{L}_\text{InfoNCE}$} &\multicolumn{4}{c}{AFHQ} & \multicolumn{4}{c}{LSUN Bedroom}\\ 
 &  	& & FID $\downarrow$ & Precision $\uparrow$ & Recall $\uparrow$ & 3D & FID $\downarrow$ & Precision $\uparrow$ & Recall $\uparrow$ & DepthFID$\downarrow$\\
\hline 
\hline
PRNeRF 			& \checkmark &				&9.14	&0.54	& 0.19	& \textcolor{blue}{Good} &14.97 	&0.55	&0.19	&110.42\\
ContraNeRF          	& &\checkmark 				&9.03 	&0.55	& 0.21	& \textcolor{blue}{Good} &15.31 	&0.54 	&0.15	&49.30\\
PR-ContraNeRF	& \checkmark & \checkmark	&9.00	&0.55	& 0.21 	& \textcolor{blue}{Good} &15.29 	&0.55 	&0.16	&55.77\\
\bottomrule
\end{tabular}}
\end{table} 
\begin{figure*}[h]
     \centering
     \begin{subfigure}[b]{0.48\linewidth}
         \centering
         \includegraphics[width=\linewidth]{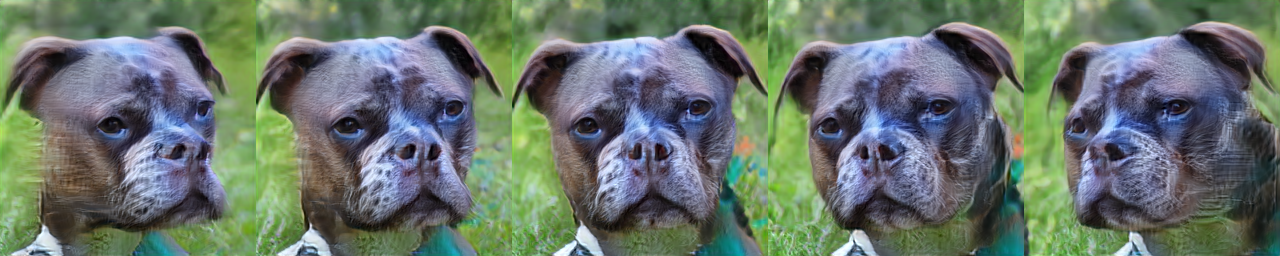}
         \includegraphics[width=\linewidth]{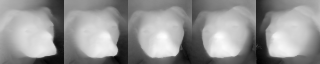}
     \end{subfigure}
     \begin{subfigure}[b]{0.48\linewidth}
         \centering
         \includegraphics[width=\linewidth]{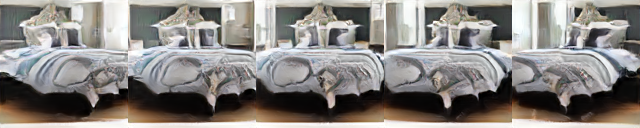}
        \includegraphics[width=\linewidth]{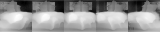}
     \end{subfigure}
     \caption{Qualitative results of the PR-ContraNeRF on the AFHQ dataset (left) and LSUN bedroom (right) dataset.
        The view angles of each scene are rotated at regular intervals: $-30^{\circ}$, $-15^{\circ}$, $0^{\circ}$, $15^{\circ}$, and $30^{\circ}$.
}
        \label{fig:prcontra_nerf}
\end{figure*}

\vspace{1cm}
\section{Implementation Details}
Our implementation is based on PyTorch~\cite{paszke2019pytorch} and employs the Adam optimizer~\cite{kingma2014adam}. 
During training, we use a batch size of 48 and apply horizontal flips for data augmentation. 
The pose embedding dimension is set to 24 for the LSUN Bedroom and AFHQ datasets, and 96 for the LSUN Church dataset. 
For CUB, we also set the pose embedding dimension to 24 and utilize available instance masks to place the birds on a white background. 
Table~\ref{tab:pose_distribution} provides additional details, including the number of images and prior pose distribution, $p_\xi$, for each dataset.

\begin{table}[h]
\centering
\caption{The number of images and prior camera pose distribution $p_\xi$ of the datasets.
}
\label{tab:pose_distribution}
\scalebox{0.9}{
\setlength\tabcolsep{7.pt}
\begin{tabular}{c|c|cc|cc}
\toprule
\multirow{2}{*}{Dataset} 	&  \multirow{2}{*}{Number of Images} 	& \multicolumn{2}{c}{Pitch} 		&   \multicolumn{2}{c}{Yaw}    \\
                  	 		&  								& Distribution & Detail 		         &  Distribution & Detail 	 \\
\hline 
\hline
LSUN Bedroom~\cite{yu2015lsun}          		&3,033,042		& Gaussian & $\mu=\pi/2$, $\sigma=0.10$ 	& Gaussian & $\mu=\pi/2$, $\sigma=0.70$\\
LSUN Church~\cite{yu2015lsun} 			&126,227			& -- & $\pi/2$							& Uniform  & $\left [ \pi/2 - 5\pi/18, \pi/2 + 5\pi/18\right ]$\\
AFHQ~\cite{choi2020stargan}				&14,630 			& Gaussian & $\mu=\pi/2$, $\sigma=0.13$	& Gaussian & $\mu=\pi/2$, $\sigma=0.19$\\
CUB~\cite{wah2011caltech}				&11,788			& Gaussian & $\mu=\pi/2$, $\sigma=0.13$	& Uniform & $\left [ \pi/2 - 3\pi/4, \pi/2 + 3\pi/4\right ]$\\
\bottomrule
\end{tabular}}
\end{table} 

\clearpage
\section{Qualitative Evaluation with High Resolution}
\begin{figure*}[h]
        \centering
         \begin{subfigure}[b]{0.99\linewidth}
         \begin{minipage}{0.5\linewidth}
         \centering
         \includegraphics[width=0.49\linewidth]{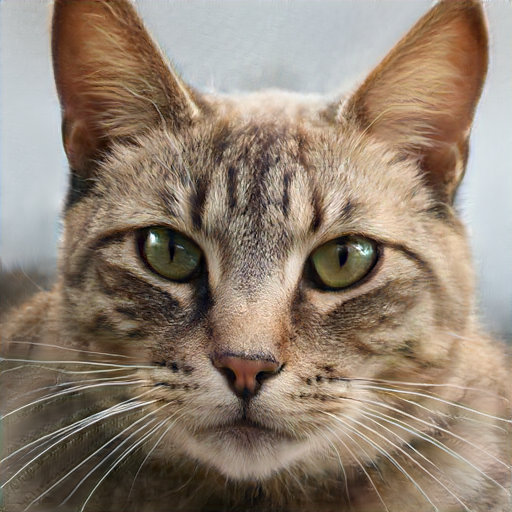}
         \includegraphics[width=0.49\linewidth]{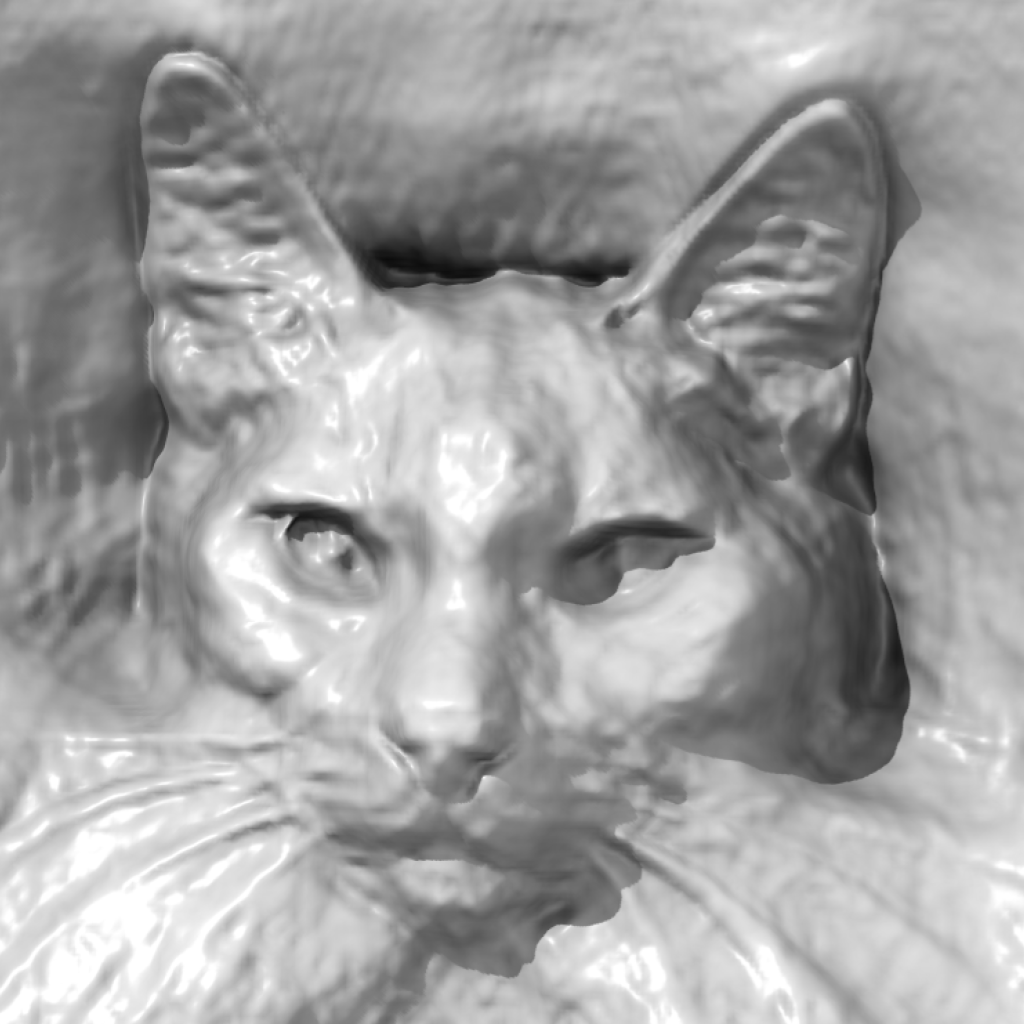}
         \end{minipage}
          \begin{minipage}{0.49\linewidth}
         \centering
         \includegraphics[width=0.99\linewidth]{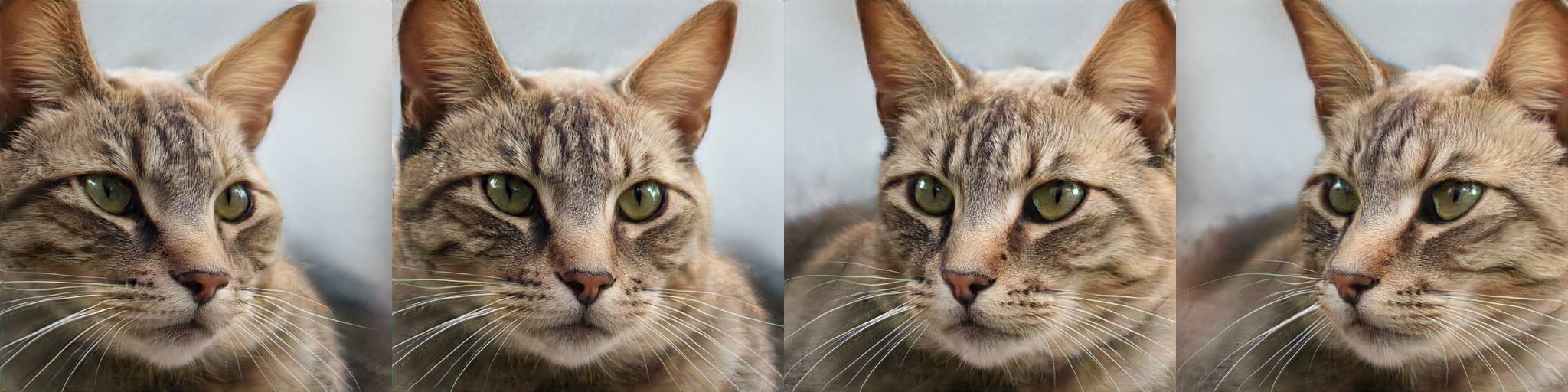}
         \includegraphics[width=0.99\linewidth]{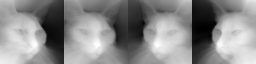}
         \end{minipage}     
     	         \vspace{1.5mm}
	\end{subfigure}	
	         \begin{subfigure}[b]{0.99\linewidth}
         \begin{minipage}{0.5\linewidth}
         \centering
         \includegraphics[width=0.49\linewidth]{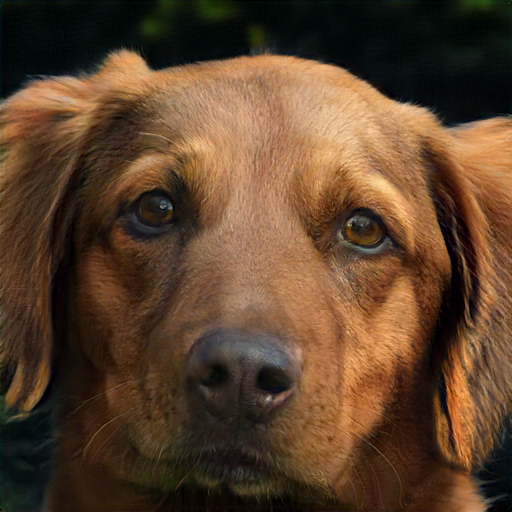}
         \includegraphics[width=0.49\linewidth]{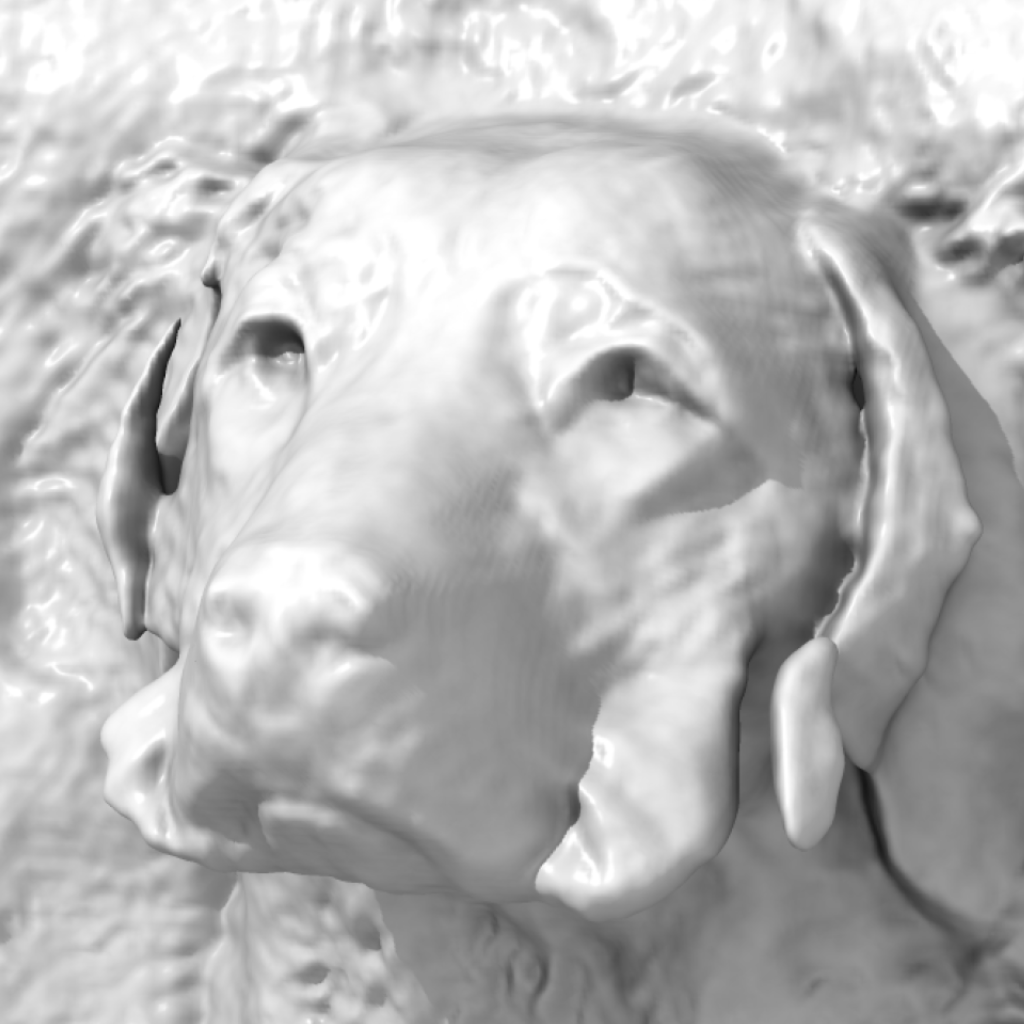}
         \end{minipage}
          \begin{minipage}{0.49\linewidth}
         \centering
         \includegraphics[width=0.99\linewidth]{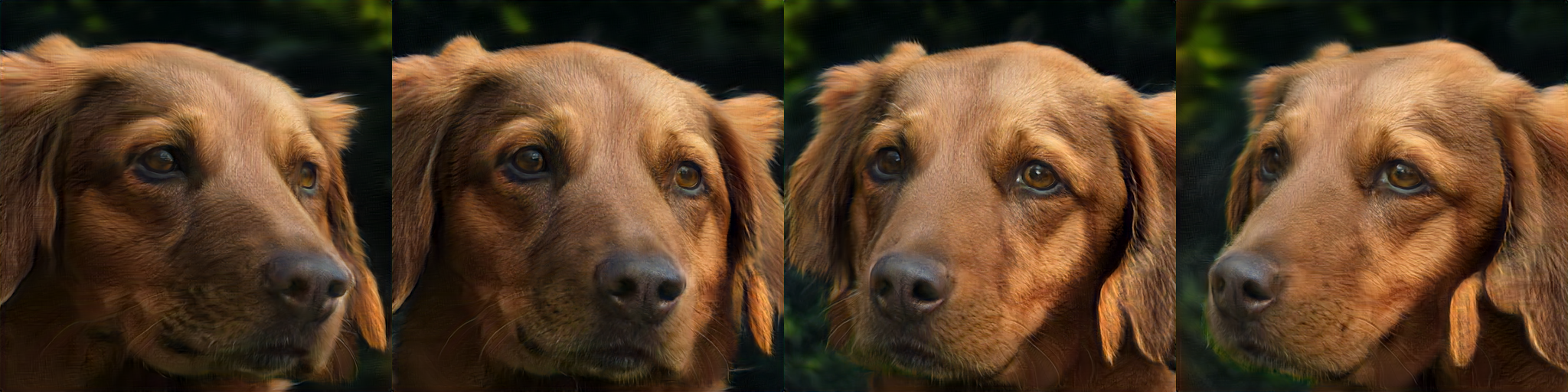}
         \includegraphics[width=0.99\linewidth]{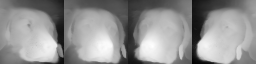}
         \end{minipage}     
     	         \vspace{1.5mm}
	\end{subfigure}	
	\begin{subfigure}[b]{0.99\linewidth}
         \begin{minipage}{0.5\linewidth}
         \centering
         \includegraphics[width=0.49\linewidth]{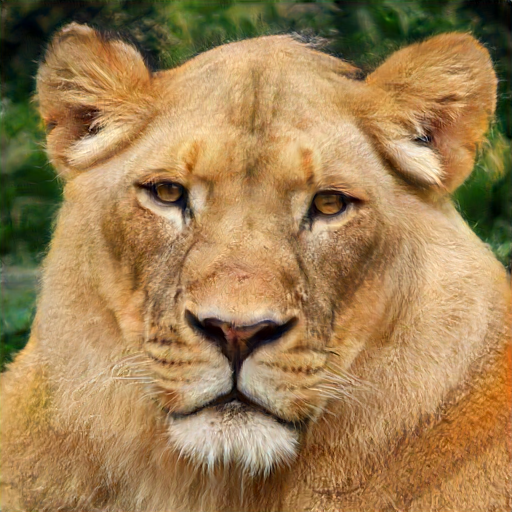}
         \includegraphics[width=0.49\linewidth]{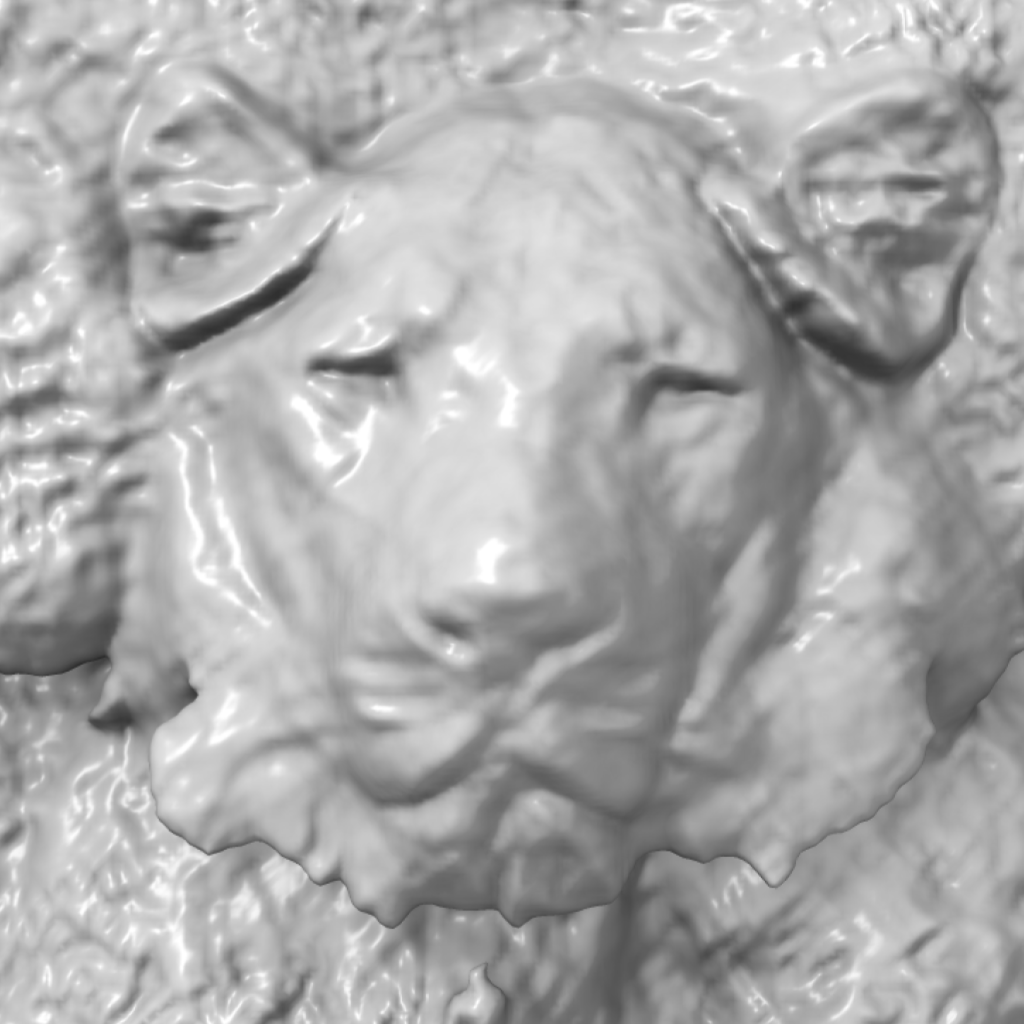}
         \end{minipage}
          \begin{minipage}{0.49\linewidth}
         \centering
         \includegraphics[width=0.99\linewidth]{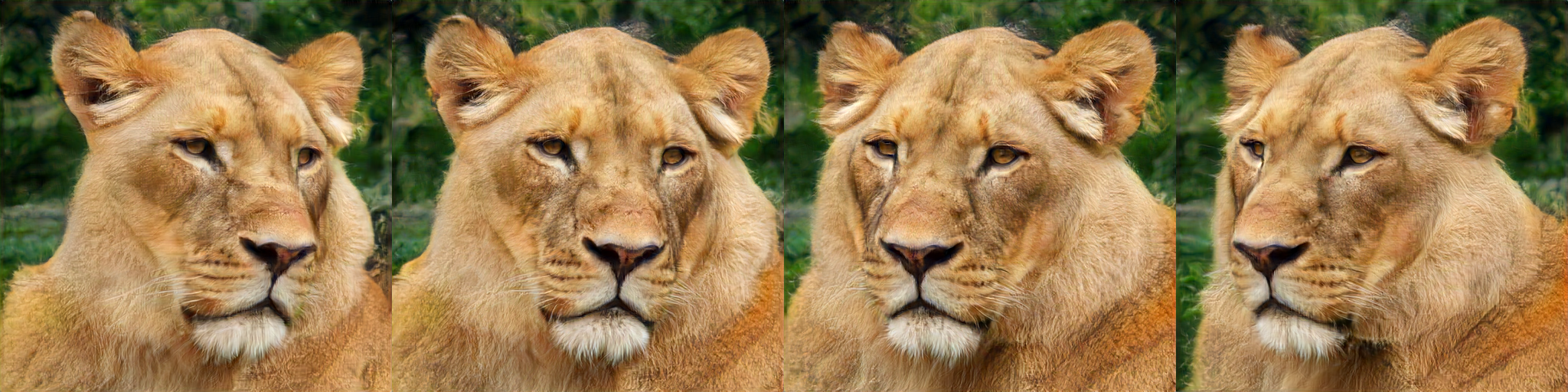}
         \includegraphics[width=0.99\linewidth]{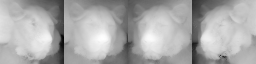}
         \end{minipage}    
         \vspace{1.5mm}
        \end{subfigure}	 
\begin{subfigure}[b]{0.99\linewidth}
         \begin{minipage}{0.5\linewidth}
         \centering
         \includegraphics[width=0.49\linewidth]{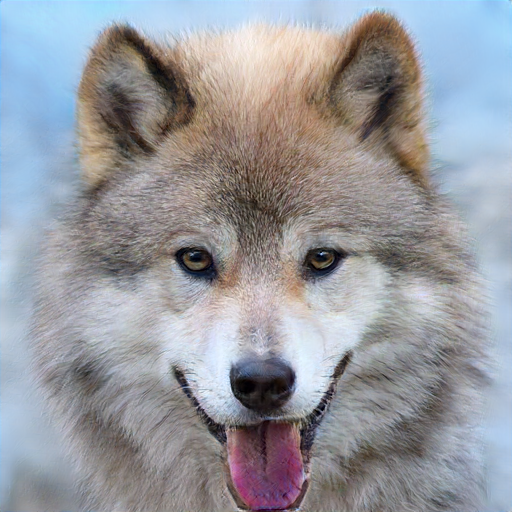}
         \includegraphics[width=0.49\linewidth]{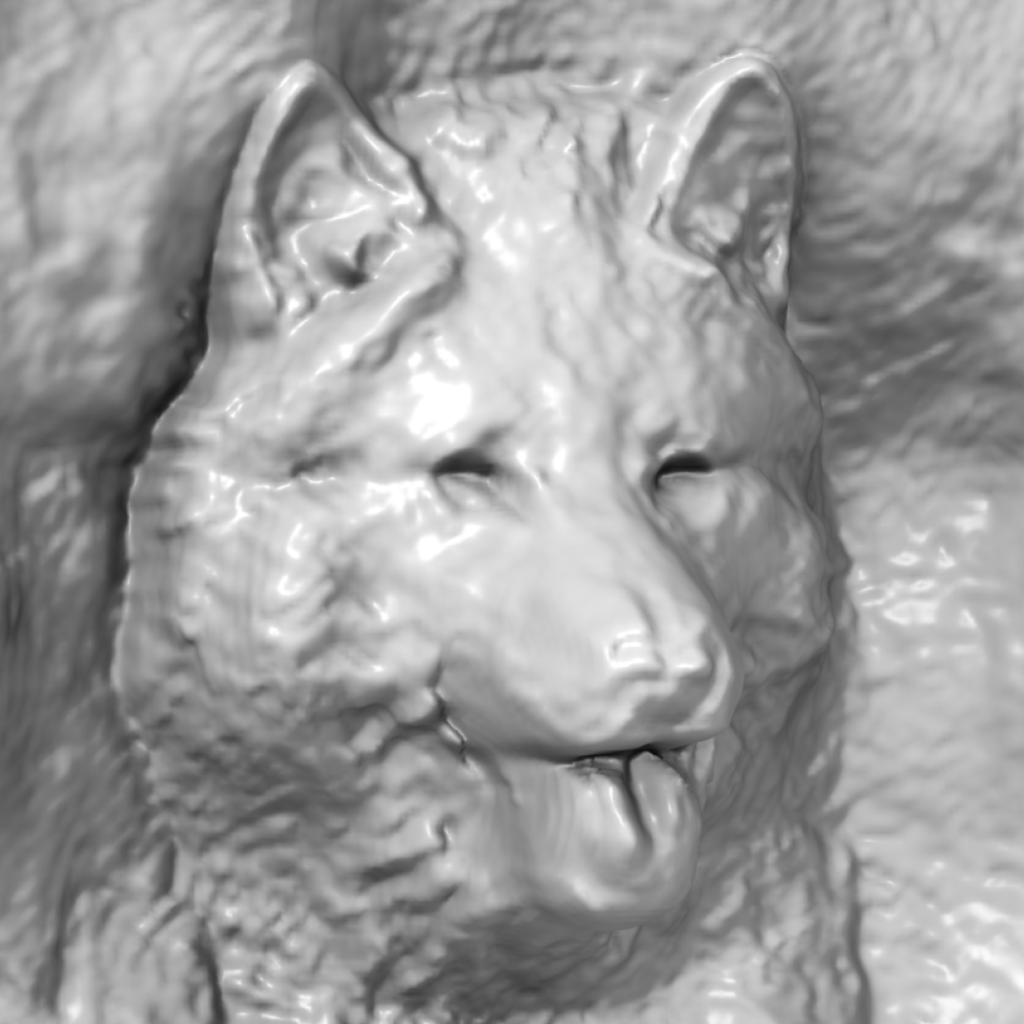}
         \end{minipage}
          \begin{minipage}{0.49\linewidth}
         \centering
         \includegraphics[width=0.99\linewidth]{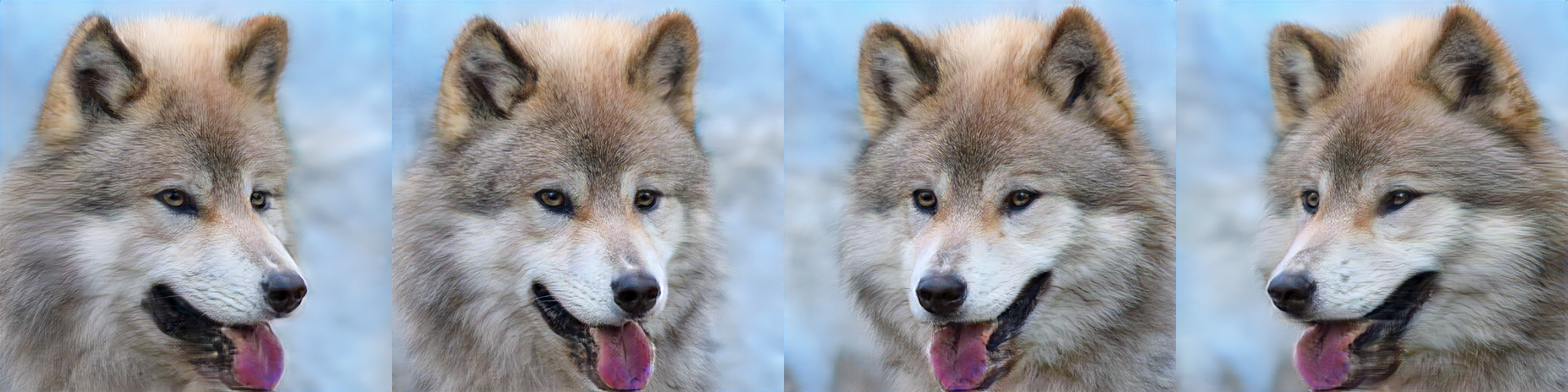}
         \includegraphics[width=0.99\linewidth]{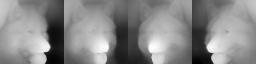}
         \end{minipage}     
     	\end{subfigure}	
		\caption{Curated examples from ContraNeRF trained on the combined AFHQ (Cats, Dogs, and Wildlifes) with $512^2$ resolution.
		The first column exhibits images rendered with the mean yaw and pitch value, and the second column shows surface renderings with random poses.
		The last column shows RGB images, and their depth maps when varying camera yaw angles at $-30^{\circ}$, $-15^{\circ}$, $15^{\circ}$, and $30^{\circ}$.
		}
		\label{fig:afhq_512}
    \end{figure*}

Figure~\ref{fig:afhq_512} exhibits the results of ContraNeRF on the AFHQ dataset, including Cats, Dogs, and Wildlifes categories with $512^2$ resolution.
ContraNeRF produces high-resolution and high-fidelity images with accurate depth maps and surface.

\clearpage
\iffalse
\section{\textcolor{red}{COLMAP Reconstruction}}
\fi
\section{Qualitative Evaluation with Vertical Rotation}
\begin{figure*}[h]
     \centering
          \begin{subfigure}[b]{0.61\linewidth}
         \centering
         \includegraphics[width=0.99\linewidth]{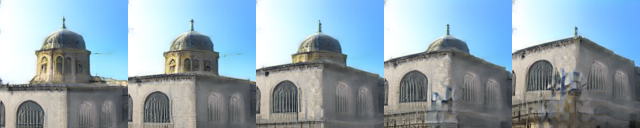}
         \includegraphics[width=0.99\linewidth]{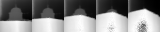}
         \includegraphics[width=0.99\linewidth]{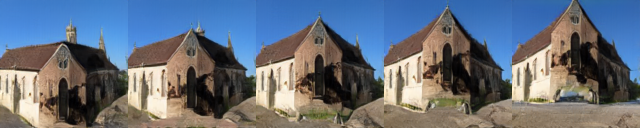}
         \includegraphics[width=0.99\linewidth]{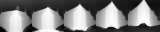}
         \caption{LSUN Church}
     \end{subfigure}
          \begin{subfigure}[b]{0.61\linewidth}
         \centering
         \includegraphics[width=0.99\linewidth]{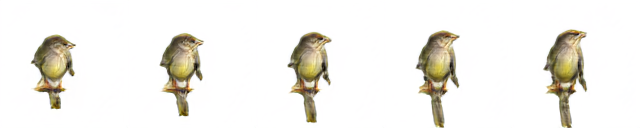}
         \includegraphics[width=0.99\linewidth]{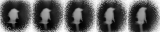}
         \includegraphics[width=0.99\linewidth]{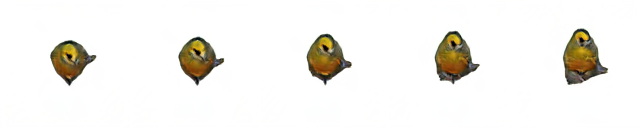}
         \includegraphics[width=0.99\linewidth]{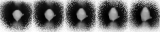}
         \caption{CUB}
     \end{subfigure}
          \caption{
       Qualitative results of ContraNeRF on the LSUN Church, and CUB dataset.
        The view angles of each scene are rotated vertically at regular intervals: $20^{\circ}$, $10^{\circ}$, $0^{\circ}$, $-10^{\circ}$, and $-20^{\circ}$.
}
        \vspace{-1cm}
        \label{fig:diverse_camera_pose1}
\end{figure*}

\begin{figure*}[h]
     \vspace{10mm}
     \centering
     \begin{subfigure}[b]{0.61\linewidth}
         \centering
	\includegraphics[width=0.99\linewidth]{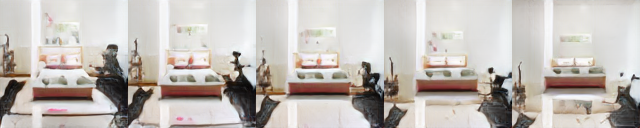}        
         \includegraphics[width=0.99\linewidth]{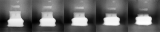}
         \includegraphics[width=0.99\linewidth]{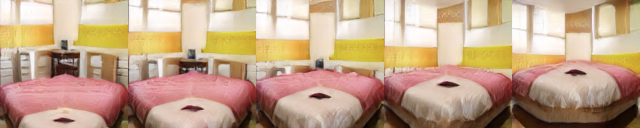}        
         \includegraphics[width=0.99\linewidth]{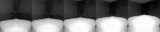}
         \caption{LSUN Bedroom}
         \vspace{2mm}
     \end{subfigure}
     \begin{subfigure}[b]{0.61\linewidth}
         \centering
         \includegraphics[width=0.99\linewidth]{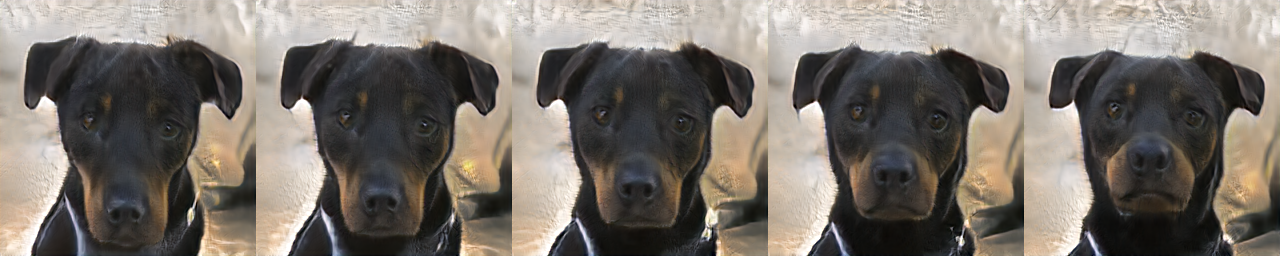}
         \includegraphics[width=0.99\linewidth]{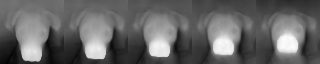}
         \includegraphics[width=0.99\linewidth]{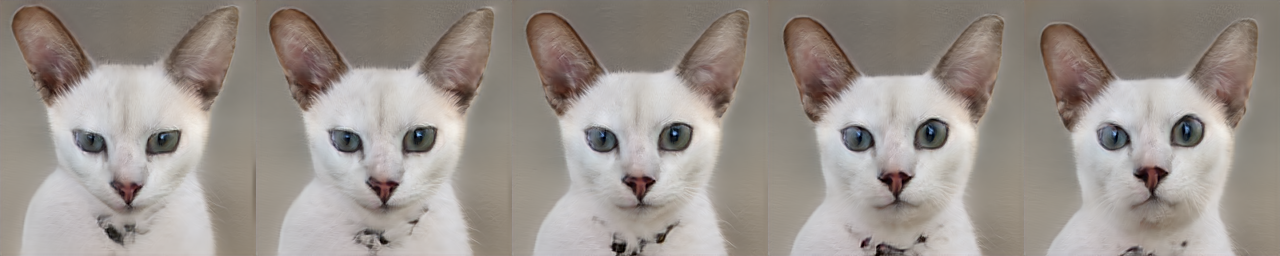}
         \includegraphics[width=0.99\linewidth]{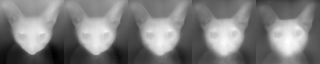}
         \caption{AFHQ dataset}
     \end{subfigure}
          \caption{
       Qualitative results of ContraNeRF on the LSUN Bedroom and AFHQ dataset.
               We visualized the rendering results with RGB image and its depth map. 
        The view angles of each scene are rotated vertically at regular intervals: $20^{\circ}$, $10^{\circ}$, $0^{\circ}$, $-10^{\circ}$, and $-20^{\circ}$.
}
        \label{fig:diverse_camera_pose2}
\end{figure*}

In the main paper, we presented our results using a camera featuring horizontal rotations. 
To further assess the effectiveness of our algorithms, particularly ContraNeRF, we display supplementary qualitative results incorporating vertical rotations, as depicted in Figure~\ref{fig:diverse_camera_pose1} and \ref{fig:diverse_camera_pose2}.
The scenes are observed from 5 distinct views, varying camera pitch angles at $20^{\circ}$, $10^{\circ}$, $0^{\circ}$, $-10^{\circ}$, and $-20^{\circ}$.
For the church dataset, the results show that ours even can generate images with a camera pose from out-of-distribution; during training, the pitch value is fixed for the church dataset.

\clearpage

\clearpage
\section{Qualitative Evaluation of 3D Reconstruction Results}

\begin{figure*}[h]
\centering
     \begin{subfigure}[b]{0.48\linewidth}
         \centering
         \includegraphics[width=\linewidth]{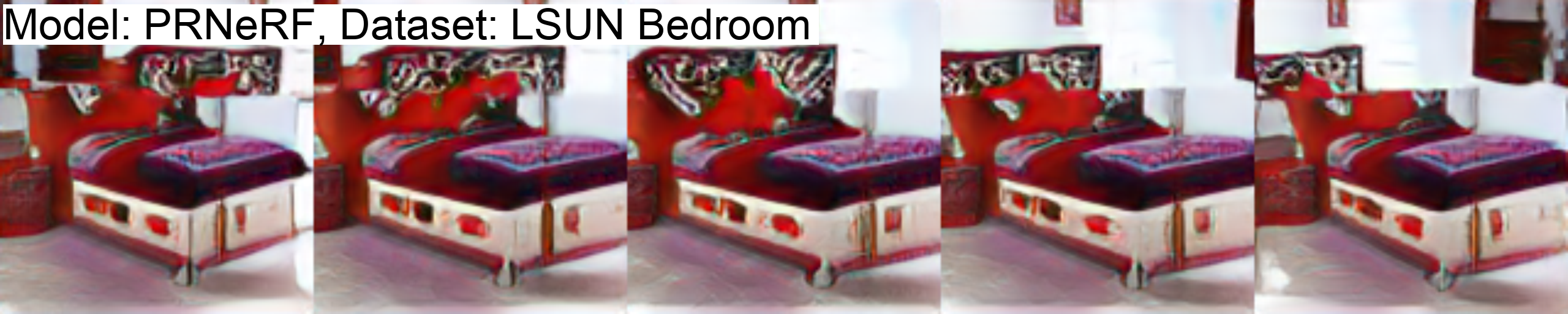}
         \includegraphics[width=\linewidth]{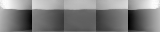}
         \includegraphics[width=\linewidth]{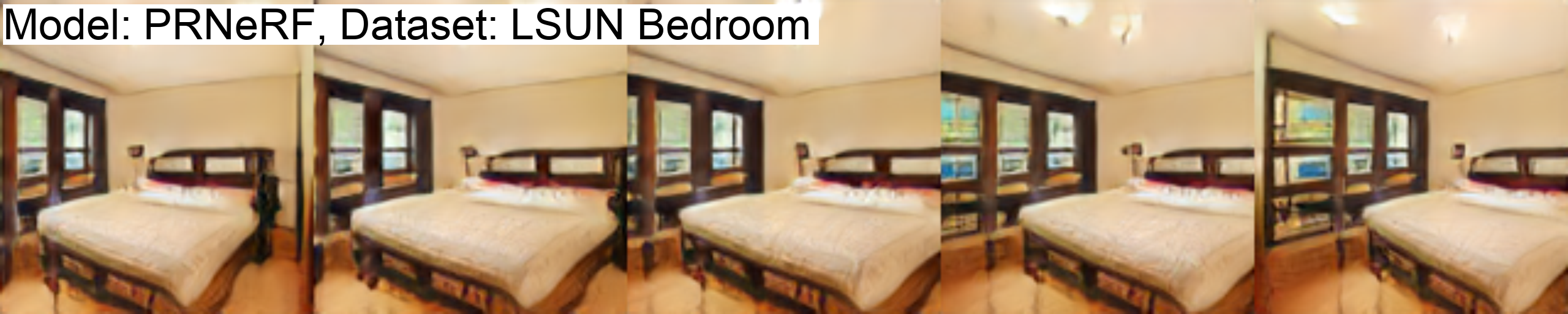}
         \includegraphics[width=\linewidth]{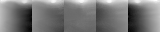}
         \includegraphics[width=\linewidth]{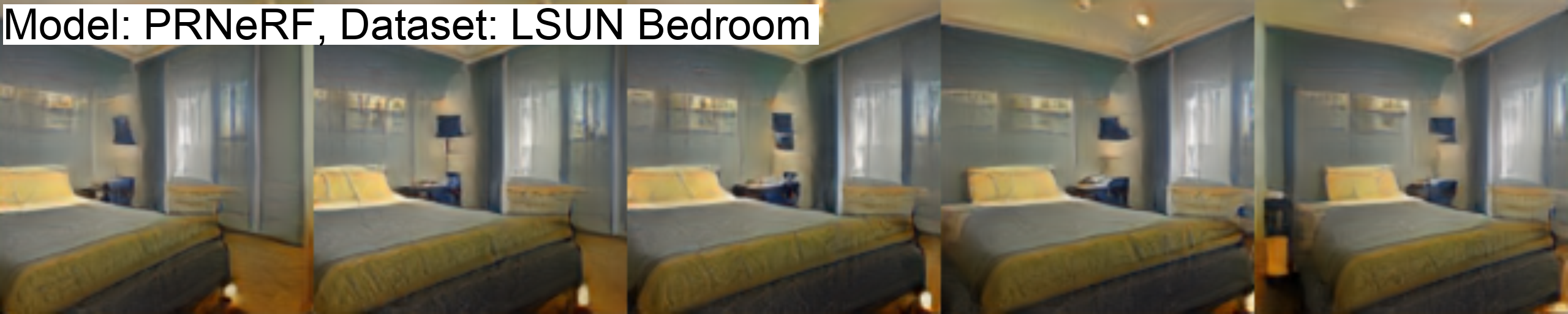}
         \includegraphics[width=\linewidth]{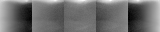}
     \end{subfigure}
     \hspace{0.1mm}
     \begin{subfigure}[b]{0.48\linewidth}
         \centering
         \includegraphics[width=\linewidth]{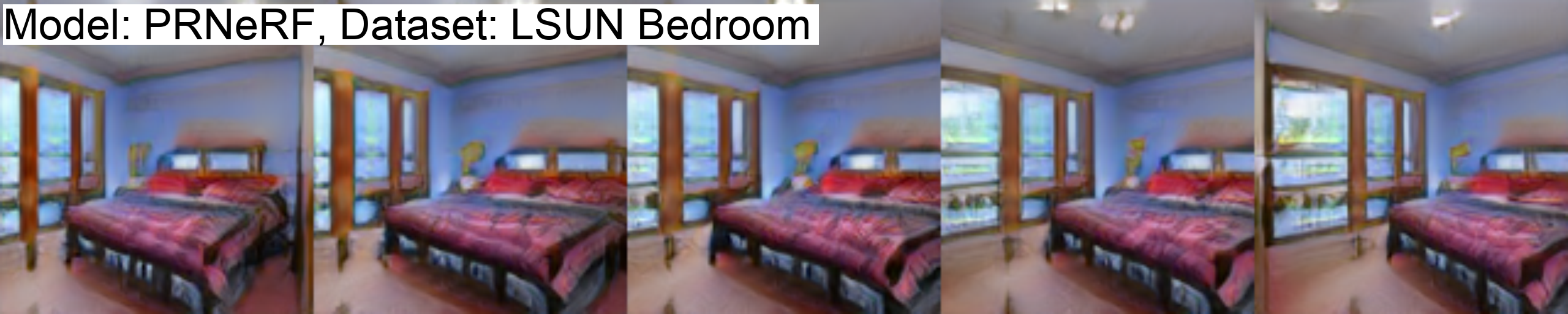}
         \includegraphics[width=\linewidth]{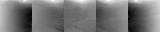}
         \includegraphics[width=\linewidth]{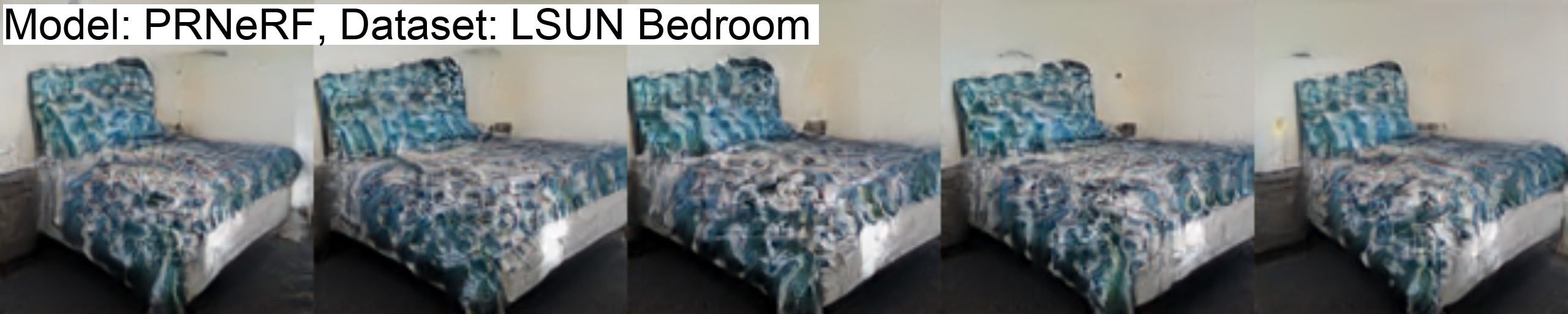}
         \includegraphics[width=\linewidth]{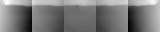}
         \includegraphics[width=\linewidth]{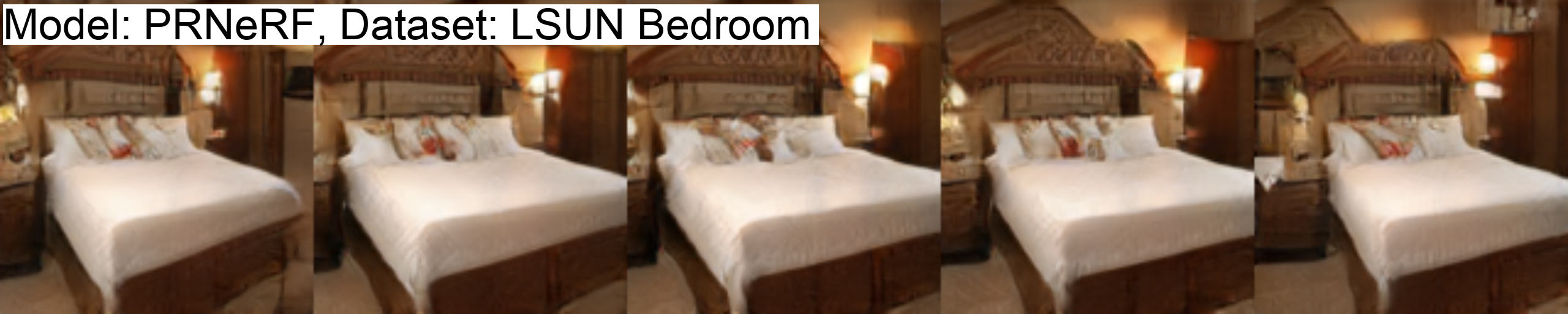}
         \includegraphics[width=\linewidth]{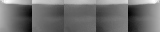}
     \end{subfigure}
     \caption{Rendered images and their \textcolor{red}{Bad} depth maps. PRNeRF, when applied to the LSUN Bedroom dataset, produces images of this quality.
     }
\label{fig:level_bad}
\end{figure*}

\begin{figure*}[h]
\vspace{-0.3cm}
\centering
     \begin{subfigure}[b]{0.48\linewidth}
         \centering
         \includegraphics[width=\linewidth]{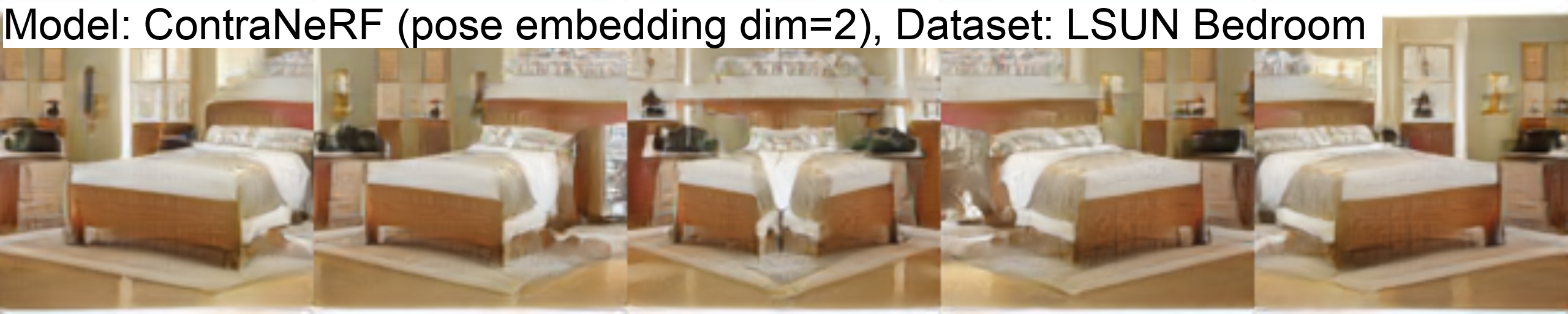}
         \includegraphics[width=\linewidth]{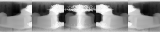}
         \includegraphics[width=\linewidth]{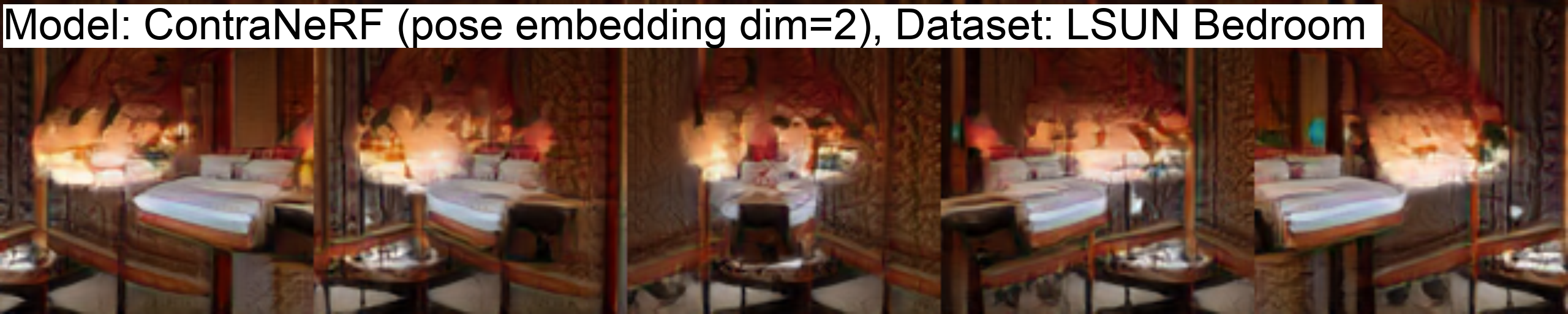}
         \includegraphics[width=\linewidth]{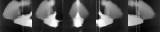}
         \includegraphics[width=\linewidth]{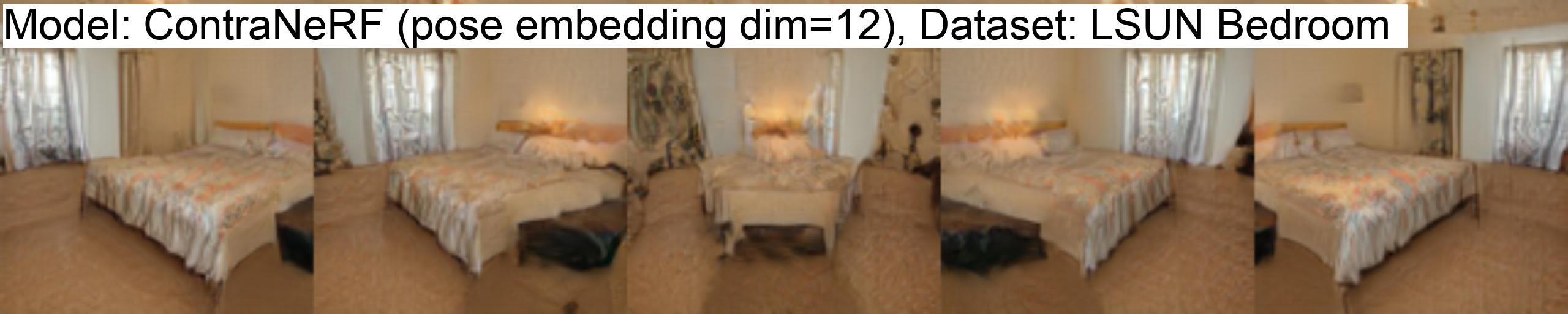}
         \includegraphics[width=\linewidth]{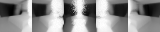}
     \end{subfigure}
     \hspace{0.1mm}
     \begin{subfigure}[b]{0.48\linewidth}
         \centering
         \includegraphics[width=\linewidth]{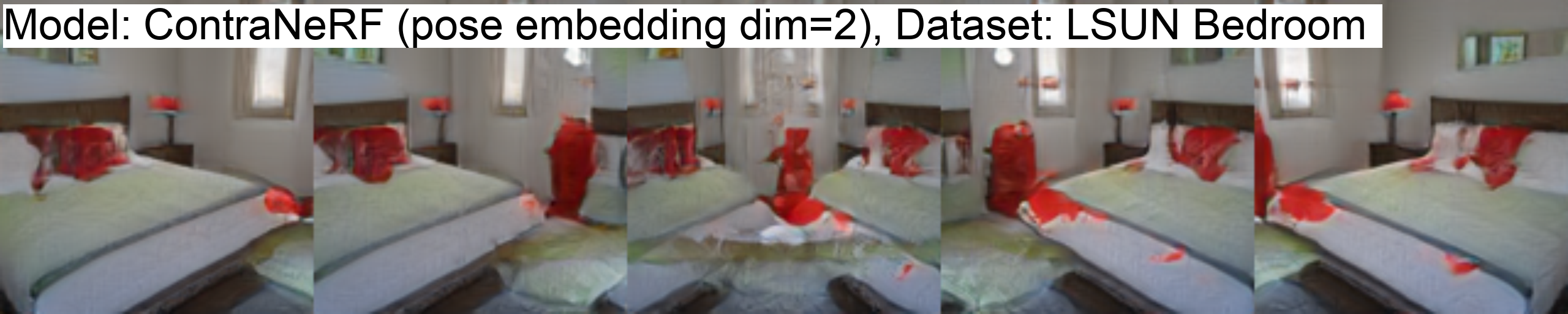}
         \includegraphics[width=\linewidth]{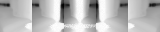}
         \includegraphics[width=\linewidth]{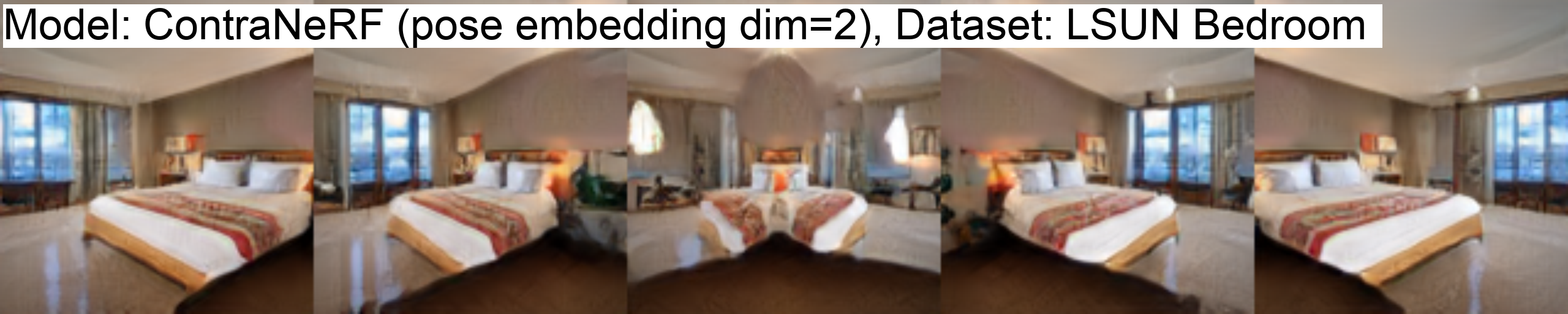}
         \includegraphics[width=\linewidth]{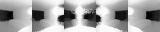}
         \includegraphics[width=\linewidth]{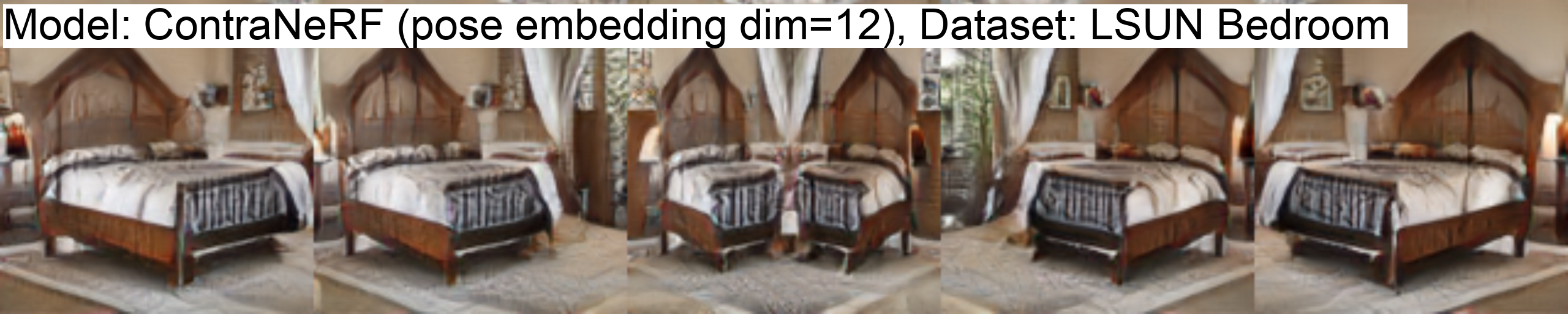}
         \includegraphics[width=\linewidth]{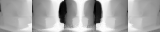}
     \end{subfigure}
	\caption{Rendered images and their \textcolor{Green}{Fair} depth maps.
	ContraNeRF with a low-dimensional camera pose embedding tends to generate images in the \textcolor{Green}{Fair} quality of depth maps on the LSUN Bedroom dataset.
	}
	\label{fig:level_fair}
\end{figure*}

\begin{figure*}[h]
\centering
     \begin{subfigure}[b]{0.48\linewidth}
         \centering
         \includegraphics[width=\linewidth]{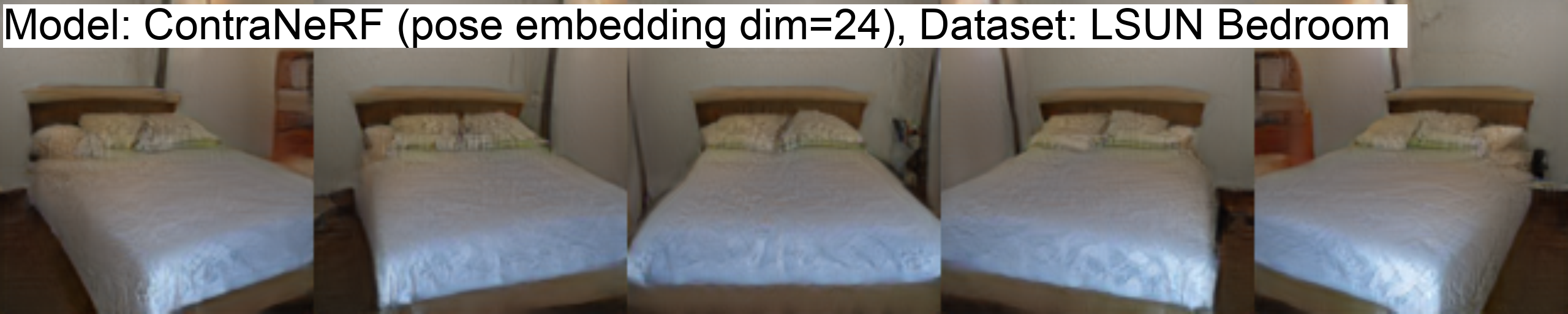}
         \includegraphics[width=\linewidth]{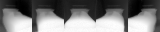}
         \includegraphics[width=\linewidth]{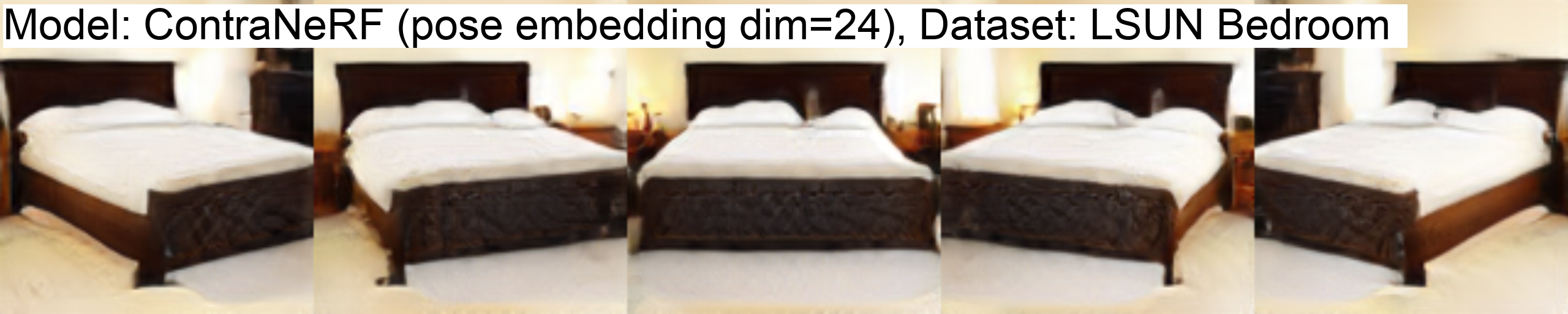}
         \includegraphics[width=\linewidth]{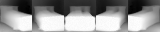}
         \includegraphics[width=\linewidth]{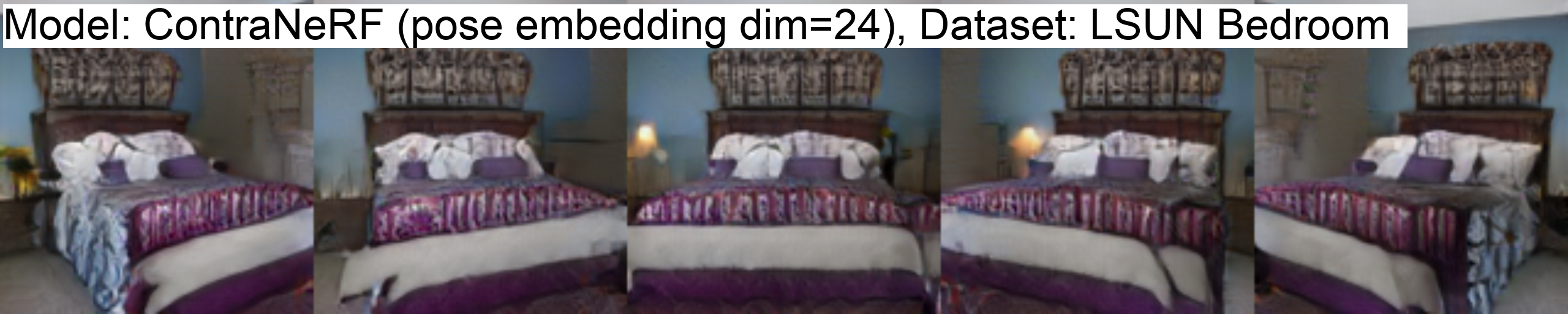}
         \includegraphics[width=\linewidth]{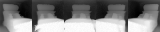}
     \end{subfigure}
     \hspace{0.1mm}
     \begin{subfigure}[b]{0.48\linewidth}
         \centering
         \includegraphics[width=\linewidth]{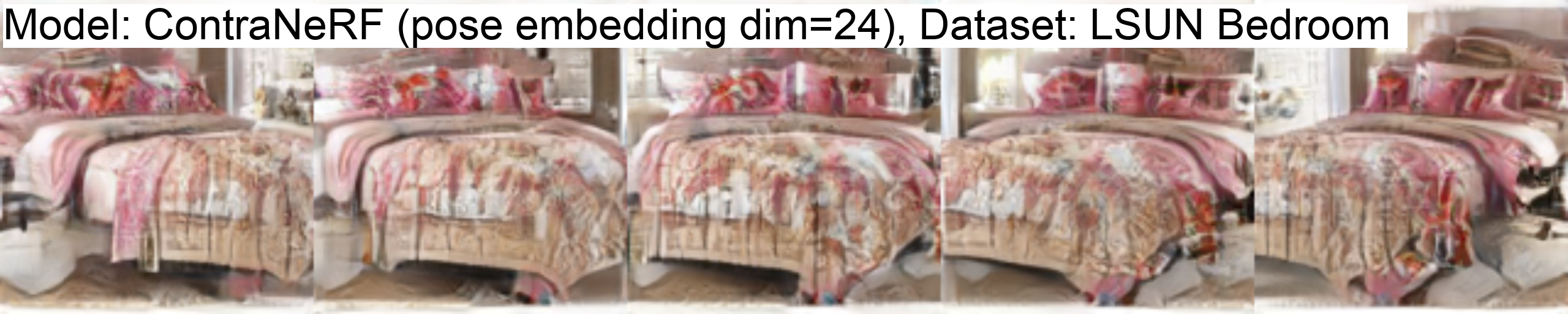}
         \includegraphics[width=\linewidth]{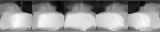}
         \includegraphics[width=\linewidth]{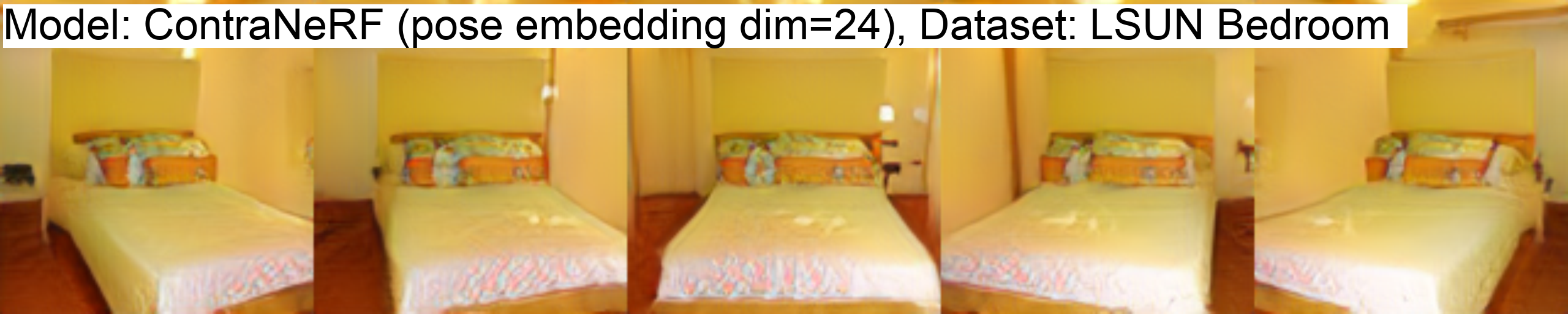}
         \includegraphics[width=\linewidth]{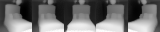}
         \includegraphics[width=\linewidth]{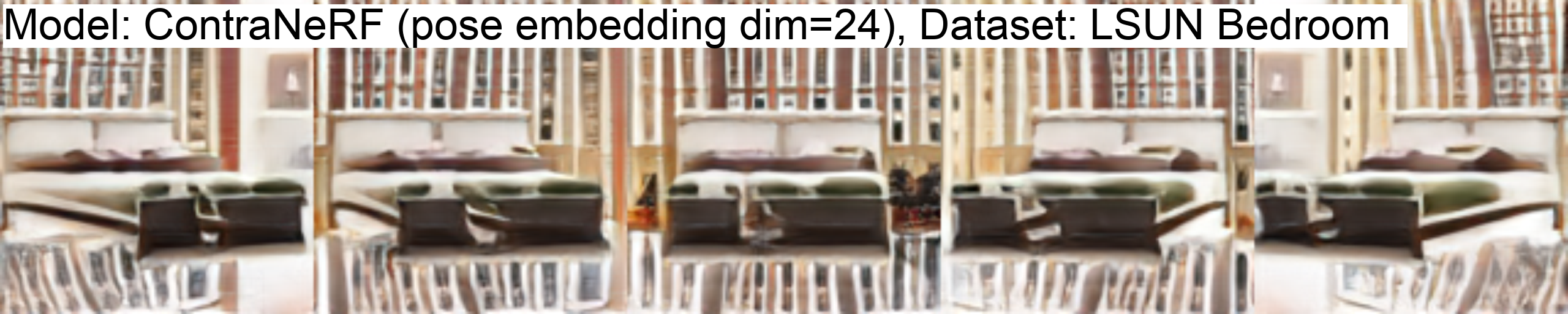}
         \includegraphics[width=\linewidth]{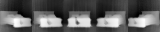}
     \end{subfigure}
	\caption{Rendered images and their \textcolor{blue}{Good} depth maps.
ContraNeRF, when a high-dimensional camera pose embedding is utilized, typically produces images with \textcolor{blue}{Good} quality depth maps.
}
\label{fig:level_good}     
\vspace{-0.5cm}
\end{figure*}

We categorize the reconstruction quality of each scene into three levels: Bad, Fair, and Good. 
Each level generally exhibits the following characteristics:

\begin{itemize}
\item \textcolor{red}{Bad}: A model is unable to discern reasonable 3D structures, resulting in planar depth maps, as demonstrated in Figure~\ref{fig:level_bad}. This quality is frequently observed in PRNeRF's outcomes on the LSUN Bedroom dataset. \vspace{-2mm}
\item \textcolor{Green}{Fair}: While a model generates volumetric scenes, the estimated depth maps contain coarse or partially inaccurate information about 3D structures, as illustrated in Figure~\ref{fig:level_fair}. ContraNeRF with low-dimensional camera pose embedding spaces (up to 12 dimensions) tends to produce this level of depth map quality. \vspace{-2mm}
\item \textcolor{blue}{Good}: A model produces high-quality depth maps that accurately represent the 3D scene structures, as shown in Figure~\ref{fig:level_good}. ContraNeRF typically attains high-quality depth maps when sufficiently high-dimensional camera pose embeddings are used.
\end{itemize}

Note that the classification of 3D scene structure quality is evident for each algorithm when applied to a specific dataset, as depicted in Figure~\ref{fig:level_bad}, \ref{fig:level_fair}, and \ref{fig:level_good}.

\vspace{0.5cm}
\vspace{-4mm}
\clearpage
\section{Additional Qualitative Results}
This section demonstrate the qualitative results of individual examples corresponding to the following four datasets: LSUN Bedroom, LSUN Church, AFHQ, and CUB.

\subsection{LSUN Bedroom}
\begin{figure*}[h]
     \centering
     \begin{subfigure}[b]{0.49\linewidth}
         \centering
         \includegraphics[width=\linewidth]{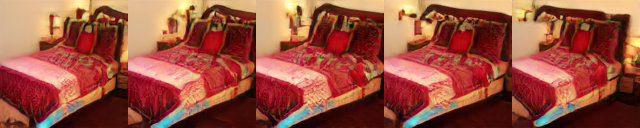}
         \includegraphics[width=\linewidth]{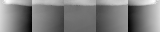}
         \includegraphics[width=\linewidth]{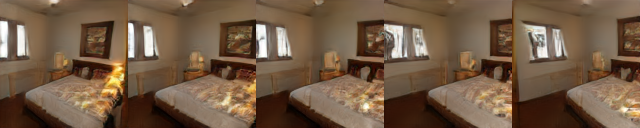}
         \includegraphics[width=\linewidth]{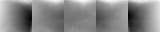}
         \includegraphics[width=\linewidth]{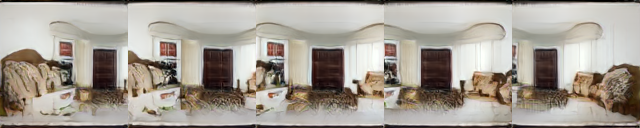}
         \includegraphics[width=\linewidth]{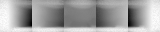}
         \includegraphics[width=\linewidth]{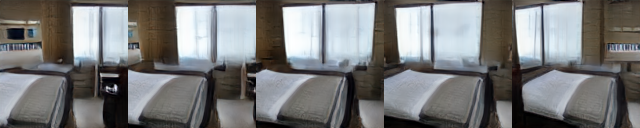}
         \includegraphics[width=\linewidth]{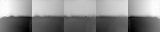}
         \includegraphics[width=\linewidth]{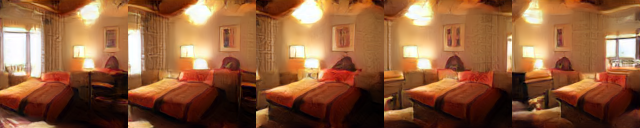}
         \includegraphics[width=\linewidth]{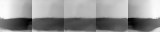}
         \caption{PRNeRF}
     \end{subfigure}
     \hfill
     \begin{subfigure}[b]{0.49\linewidth}
         \centering
         \includegraphics[width=\linewidth]{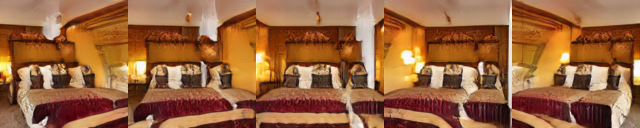}
         \includegraphics[width=\linewidth]{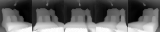}
         \includegraphics[width=\linewidth]{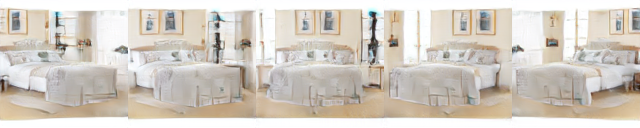}
         \includegraphics[width=\linewidth]{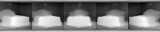}
         \includegraphics[width=\linewidth]{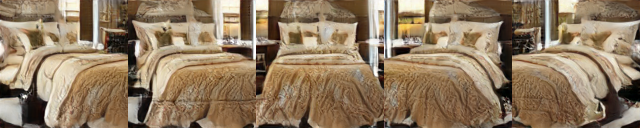}
         \includegraphics[width=\linewidth]{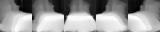}
         \includegraphics[width=\linewidth]{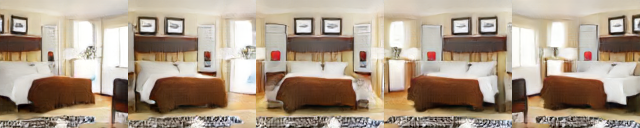}
         \includegraphics[width=\linewidth]{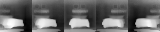}
         \includegraphics[width=\linewidth]{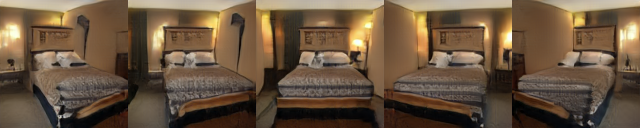}
         \includegraphics[width=\linewidth]{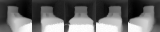}
          \caption{ContraNeRF}
     \end{subfigure}
        \caption{
        Qualitative results of PRNeRF and ContraNeRF on the LSUN Bedroom dataset.
        We visualized the rendering results with RGB image and its depth map. 
        The view angles of each scene are rotated at regular intervals: $-40^{\circ}$, $-20^{\circ}$, $0^{\circ}$, $20^{\circ}$, and $40^{\circ}$.
        }
        \label{fig:supple_qualitative_bedroom}
\end{figure*}
\newpage
\subsection{LSUN Church}
\begin{figure*}[h]
     \centering
     \begin{subfigure}[b]{0.49\linewidth}
         \centering
         \includegraphics[width=\linewidth]{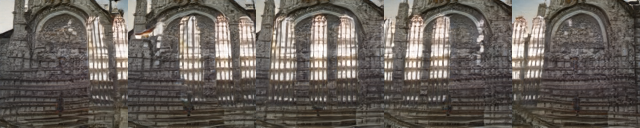}
         \includegraphics[width=\linewidth]{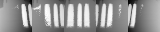}
         \includegraphics[width=\linewidth]{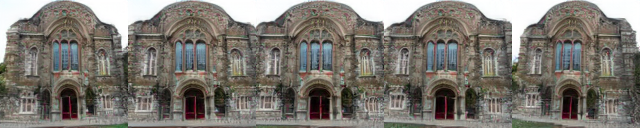}
         \includegraphics[width=\linewidth]{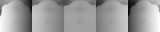}
         \includegraphics[width=\linewidth]{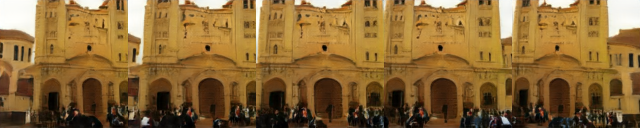}
         \includegraphics[width=\linewidth]{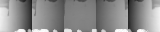}
         \includegraphics[width=\linewidth]{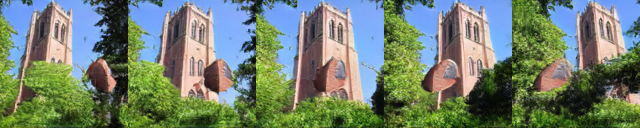}
         \includegraphics[width=\linewidth]{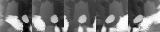}
         \includegraphics[width=\linewidth]{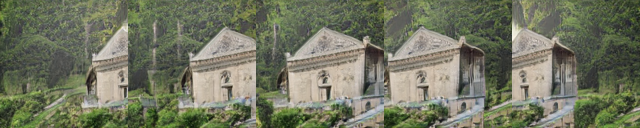}
         \includegraphics[width=\linewidth]{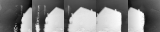}
         \caption{PRNeRF}
     \end{subfigure}
     \hfill
     \begin{subfigure}[b]{0.49\linewidth}
         \centering
         \includegraphics[width=\linewidth]{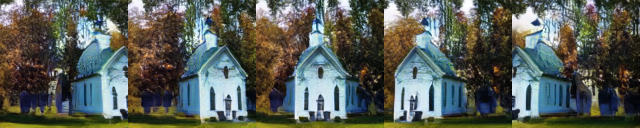}
         \includegraphics[width=\linewidth]{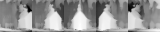}      
          \includegraphics[width=\linewidth]{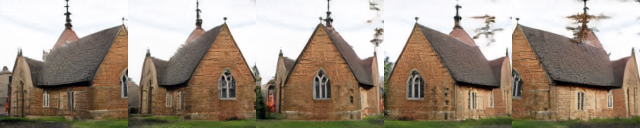}
         \includegraphics[width=\linewidth]{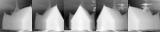}
         \includegraphics[width=\linewidth]{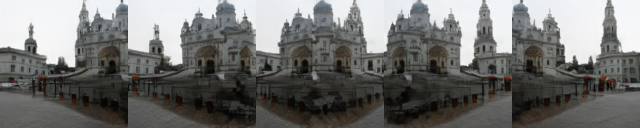}
         \includegraphics[width=\linewidth]{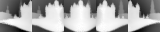}
         \includegraphics[width=\linewidth]{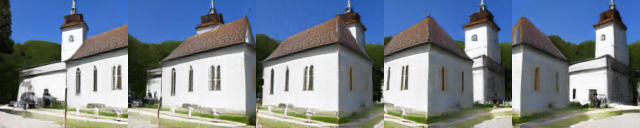}
         \includegraphics[width=\linewidth]{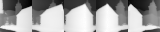}
         \includegraphics[width=\linewidth]{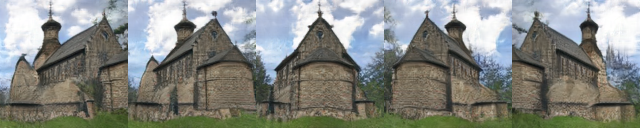}
         \includegraphics[width=\linewidth]{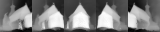}
         \caption{ContraNeRF}
     \end{subfigure}
        \caption{
        Qualitative results of PRNeRF and ContraNeRF on the LSUN Church dataset.
        We visualized the rendering results with RGB image and its depth map. 
        The view angles of each scene are rotated at regular intervals: $-40^{\circ}$, $-20^{\circ}$, $0^{\circ}$, $20^{\circ}$, and $40^{\circ}$.
       }
        \label{fig:supple_qualitative_church}
\end{figure*}
\newpage
\subsection{AFHQ}
\begin{figure*}[h]
     \centering
     \begin{subfigure}[b]{0.49\linewidth}
         \centering
         \includegraphics[width=\linewidth]{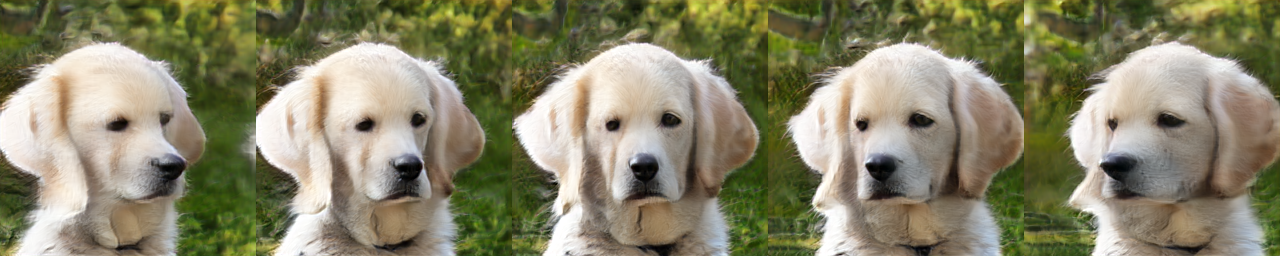}
         \includegraphics[width=\linewidth]{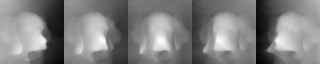}
          \includegraphics[width=\linewidth]{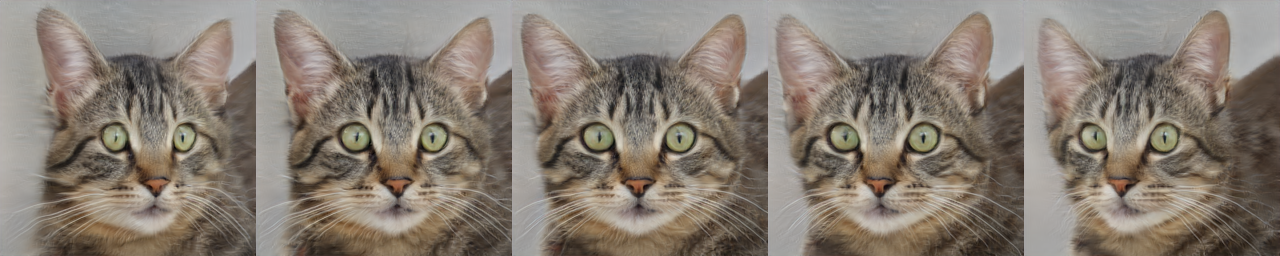}
         \includegraphics[width=\linewidth]{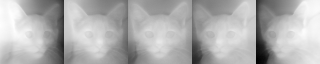}
          \includegraphics[width=\linewidth]{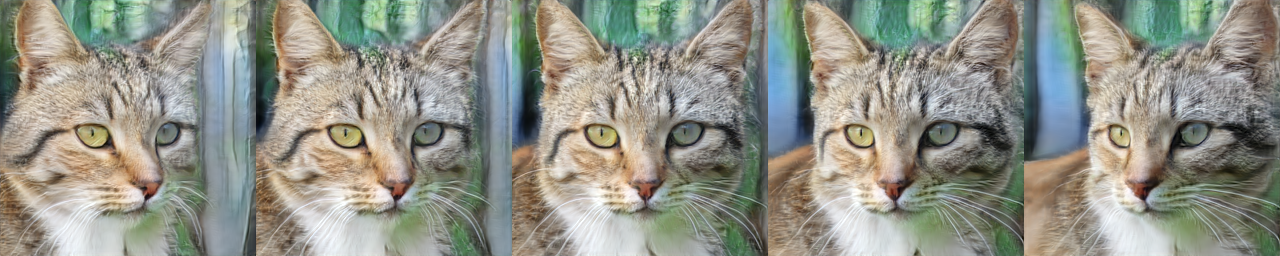}
         \includegraphics[width=\linewidth]{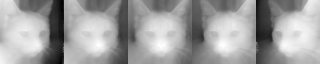}
          \includegraphics[width=\linewidth]{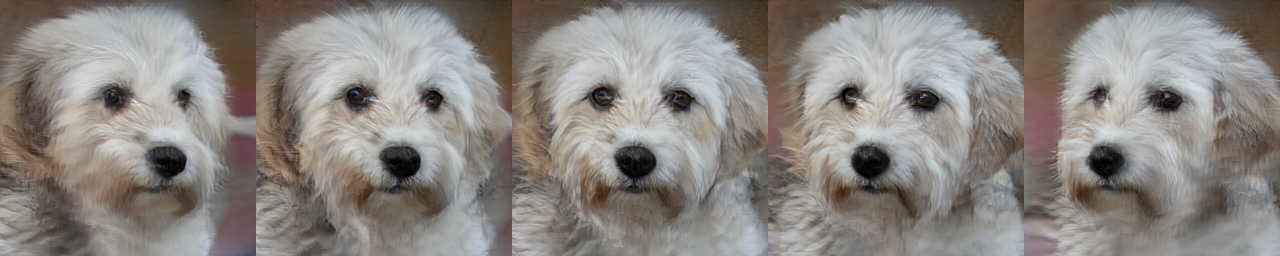}
         \includegraphics[width=\linewidth]{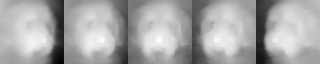}
          \includegraphics[width=\linewidth]{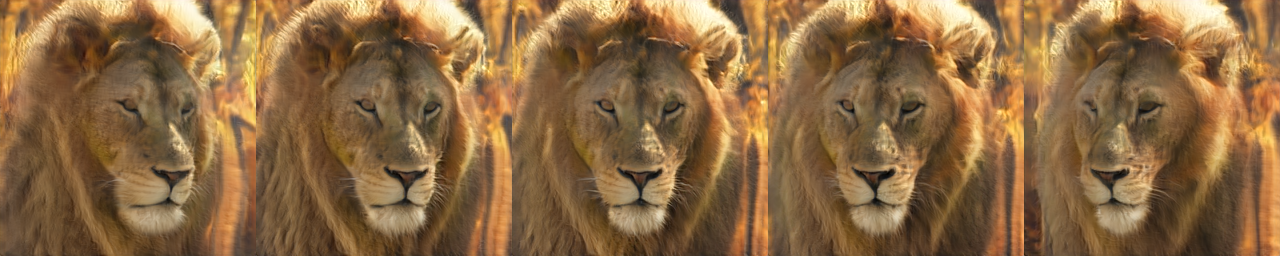}
         \includegraphics[width=\linewidth]{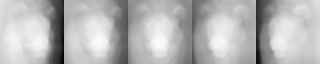}
         \caption{PRNeRF}    
     \end{subfigure}
     \hfill
     \begin{subfigure}[b]{0.49\linewidth}
         \centering
         \includegraphics[width=\linewidth]{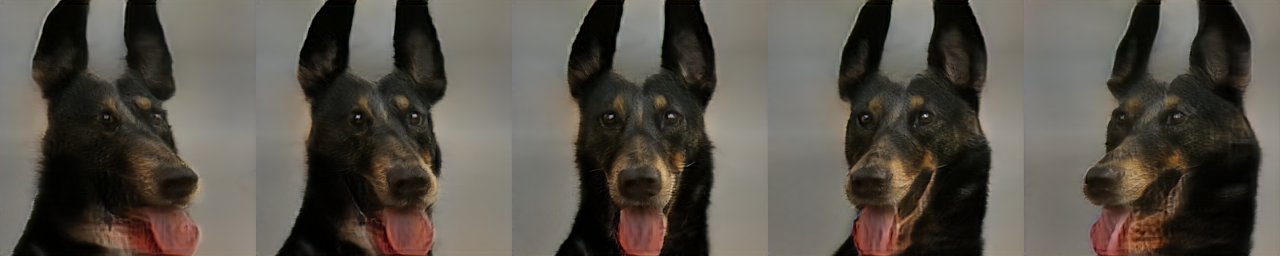}
         \includegraphics[width=\linewidth]{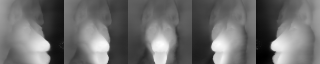}
         \includegraphics[width=\linewidth]{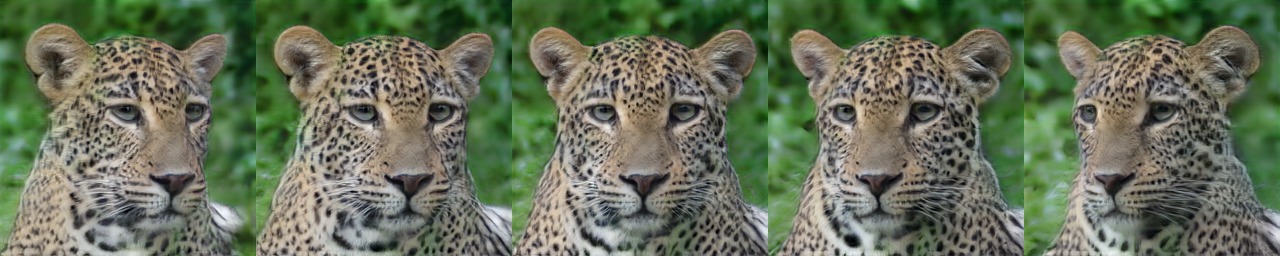}
         \includegraphics[width=\linewidth]{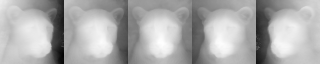}
          \includegraphics[width=\linewidth]{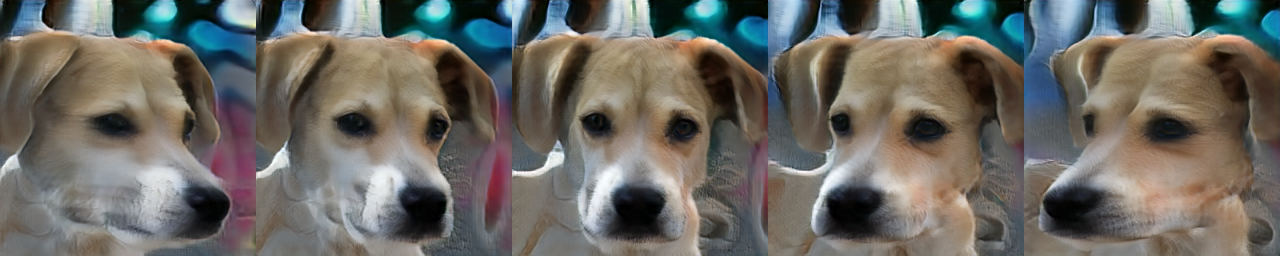}
         \includegraphics[width=\linewidth]{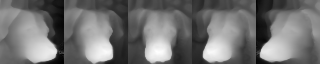}
         \includegraphics[width=\linewidth]{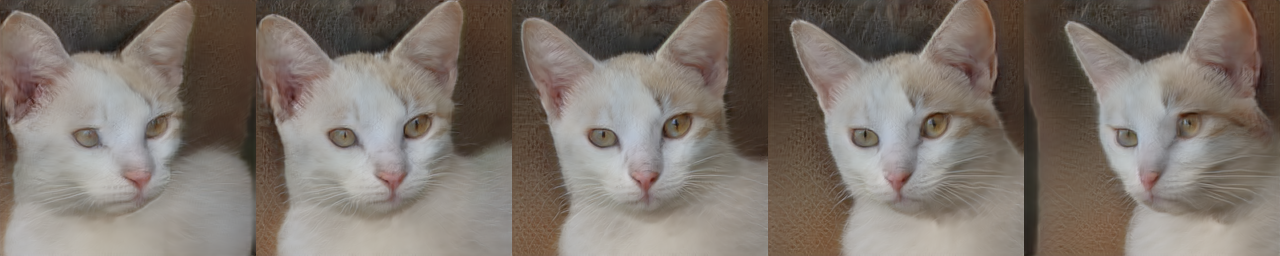}
         \includegraphics[width=\linewidth]{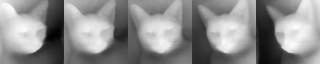}
          \includegraphics[width=\linewidth]{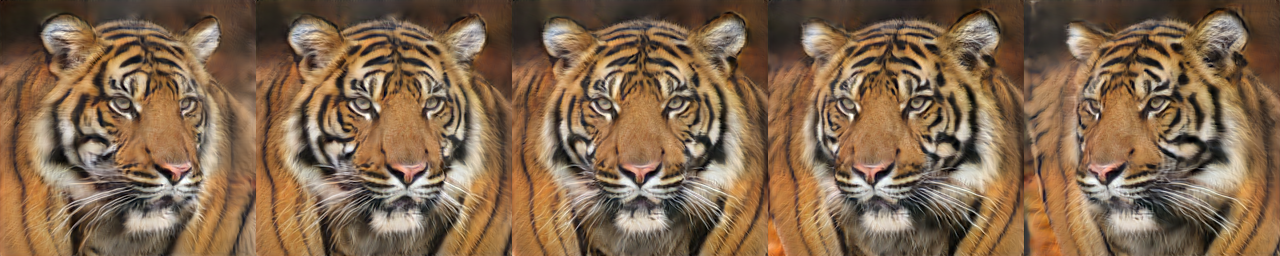}
         \includegraphics[width=\linewidth]{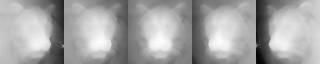}
         \caption{ContraNeRF}
     \end{subfigure}
        \caption{
        Qualitative results of PRNeRF and ContraNeRF on the AFHQ dataset.
        We visualized the rendering results with RGB image and its depth map. 
        The view angles of each scene are rotated at regular intervals: $-30^{\circ}$, $-15^{\circ}$, $0^{\circ}$, $15^{\circ}$, and $30^{\circ}$.
        }
        \label{fig:supple_qualitative_afhq}
\end{figure*}

\newpage
\subsection{CUB}
\begin{figure*}[h]
     \centering
     \begin{subfigure}[b]{0.49\linewidth}
         \centering
         \includegraphics[width=\linewidth]{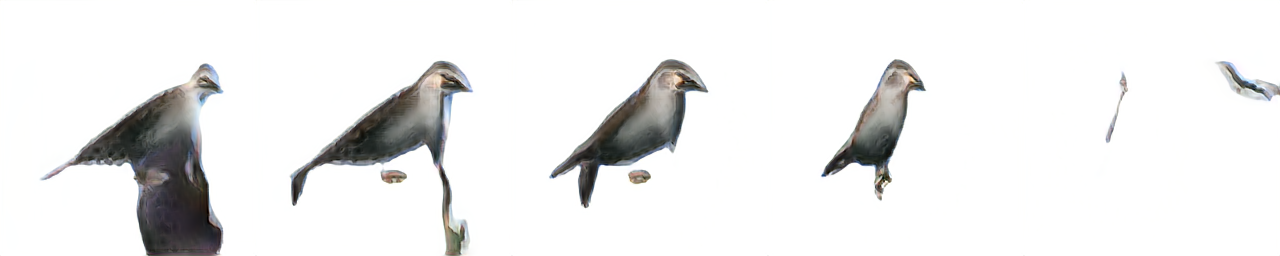}
         \includegraphics[width=\linewidth]{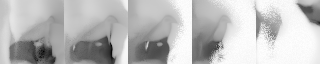}
         \includegraphics[width=\linewidth]{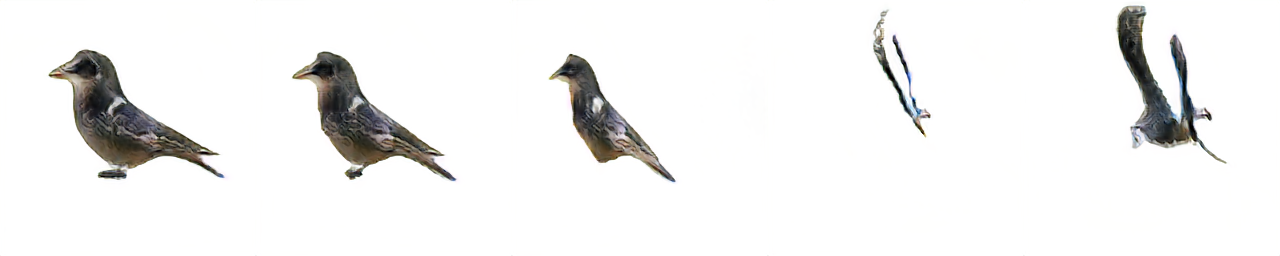}
         \includegraphics[width=\linewidth]{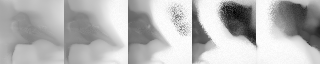}
         \includegraphics[width=\linewidth]{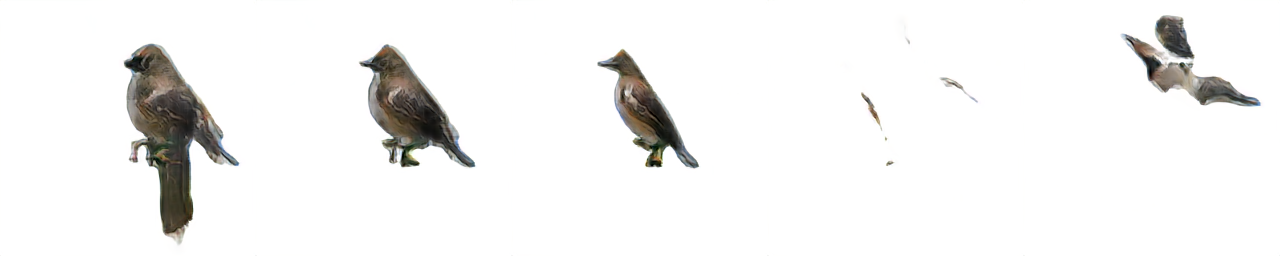}
         \includegraphics[width=\linewidth]{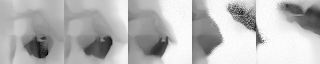}
         \includegraphics[width=\linewidth]{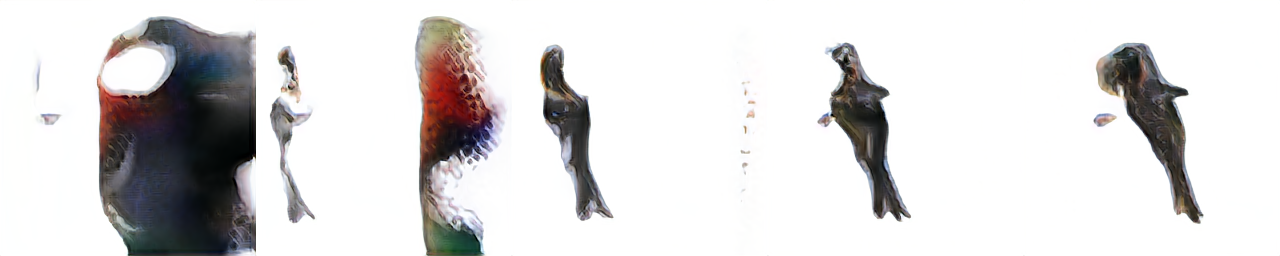}
         \includegraphics[width=\linewidth]{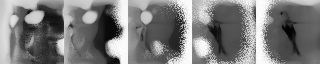}
         \includegraphics[width=\linewidth]{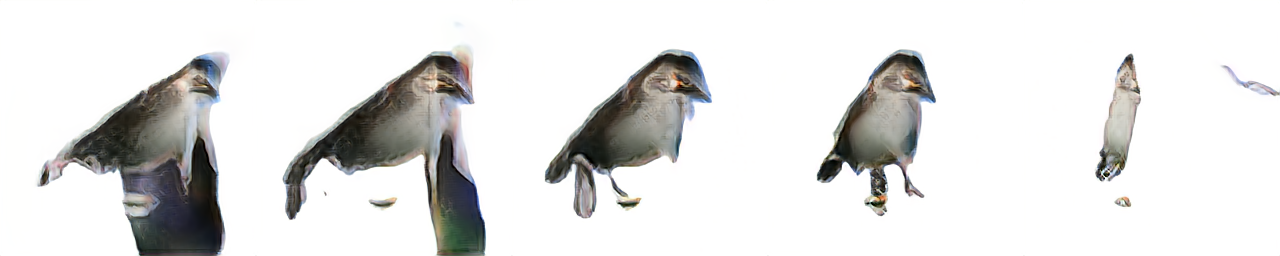}
         \includegraphics[width=\linewidth]{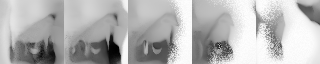}
         \caption{PRNeRF}    
     \end{subfigure}
     \hfill
     \begin{subfigure}[b]{0.49\linewidth}
         \centering
         \includegraphics[width=\linewidth]{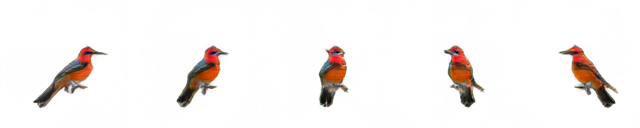}
         \includegraphics[width=\linewidth]{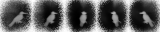}
         \includegraphics[width=\linewidth]{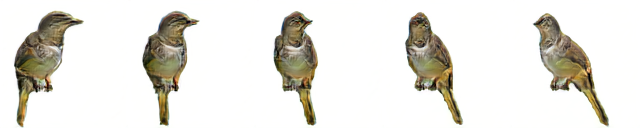}
         \includegraphics[width=\linewidth]{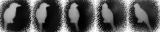}
         \includegraphics[width=\linewidth]{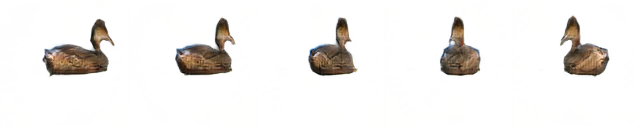}
         \includegraphics[width=\linewidth]{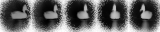}
         \includegraphics[width=\linewidth]{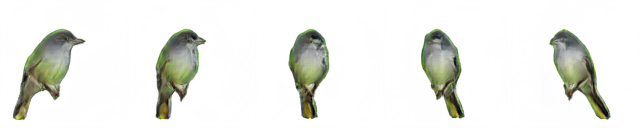}
         \includegraphics[width=\linewidth]{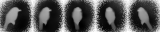}
         \includegraphics[width=\linewidth]{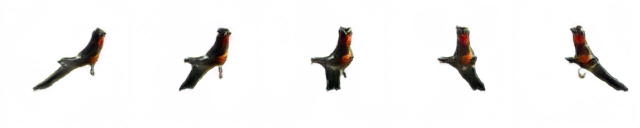}
         \includegraphics[width=\linewidth]{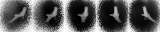}
         \caption{ContraNeRF}
     \end{subfigure}
        \caption{
        Qualitative results of PRNeRF and ContraNeRF on the CUB dataset.
        We visualized the rendering results with RGB image and its depth map. 
        The view angles of each scene are rotated at regular intervals: $-40^{\circ}$, $-20^{\circ}$, $0^{\circ}$, $20^{\circ}$, and $40^{\circ}$.
        }
        \label{fig:supple_qualitative_cub}
\end{figure*}